\documentclass{article}

\usepackage[preprint]{neurips_2023}

\usepackage[utf8]{inputenc} 
\usepackage[T1]{fontenc}    
\usepackage{url}            
\usepackage{booktabs}       
\usepackage{amsfonts}       
\usepackage{nicefrac}       
\usepackage{microtype}      
\usepackage{xcolor}         
\usepackage{graphicx}
\usepackage{subfigure}
\usepackage{hyperref}
\usepackage{csquotes}
\usepackage{multirow}
\usepackage{amsmath}
\usepackage{wrapfig}
\usepackage{algpseudocode}
\usepackage[ruled,linesnumbered]{algorithm2e}

\title{\raisebox{-0.10cm}{\includegraphics[scale=0.08]{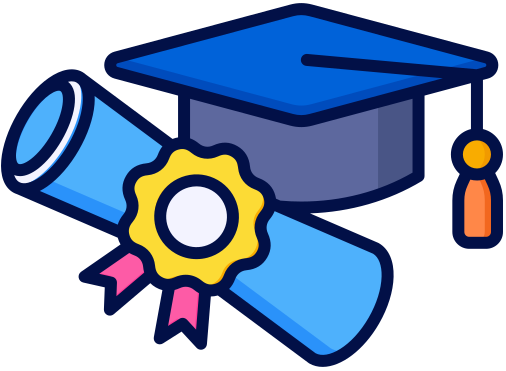}} \textcolor[rgb]{0.965,0.325,0.078}{Open}\textcolor[rgb]{0,0.631,0.945}{B}\textcolor[rgb]{1,0.733,0}{A}-V2: Reaching 77.3\% High Compression Ratio with Fast Multi-Stage Pruning}

\author{Dan Qiao\thanks{\; Equal Contribution.},
Yi Su\footnotemark[1],
Pinzheng Wang\footnotemark[1],
Jing Ye, WenJing Xie, Yuechi Zhou, Yuyang Ding, 
\AND
Zecheng Tang, Jikai Wang, Yixin Ji, Yue Wang, Pei Guo, Zechen Sun, Zikang Zhang, 
Juntao Li\thanks{\; Corresponding author. ljt@suda.edu.cn}, 
\AND
Pingfu Chao, Wenliang Chen, Guohong Fu, Guodong Zhou, Qiaoming Zhu, Min Zhang
\AND
Soochow University
}

\begin{document}
\maketitle

\begin{abstract}
Large Language Models (LLMs) have played an important role in many fields due to their powerful capabilities.
However, their massive number of parameters leads to high deployment requirements and incurs significant inference costs, which impedes their practical applications.
Training smaller models is an effective way to address this problem.
Therefore, we introduce OpenBA-V2, a 3.4B model derived from multi-stage compression and continual pre-training from the original 15B OpenBA model.
OpenBA-V2 utilizes more data, more flexible training objectives, and techniques such as layer pruning, neural pruning, and vocabulary pruning to achieve a compression rate of 77.3\% with minimal performance loss.
OpenBA-V2 demonstrates competitive performance compared to other open-source models of similar size, achieving results close to or on par with the 15B OpenBA model in downstream tasks such as common sense reasoning and Named Entity Recognition (NER).
OpenBA-V2 illustrates that LLMs can be compressed into smaller ones with minimal performance loss by employing advanced training objectives and data strategies, which may help deploy LLMs in resource-limited scenarios.
\end{abstract}

\section{Introduction}




In recent years, Large Language Models (LLMs) have demonstrated powerful capabilities in natural language understanding and generation, leading to significant achievements in various tasks such as dialogue generation, code generation, text summarization, and machine translation \citep{OpenAI2023GPT4TR,llama2,jiang2023mistral,bai2023qwen,li2023openba}.
However, their extensive demand for computing resources makes them impractical in resource-limited scenarios, such as PCs or mobile phones \citep{thawakar2024mobillama}. 
Furthermore, the high costs of inference and storage impede their widespread application across various industries \citep{bai2024beyond}.

To address these challenges, many researchers attempt to reduce the computational and storage requirements of LLMs by designing smaller models.
These smaller models are usually obtained by training from scratch \citep{openllama,zhang2024tinyllama,li2023textbooks,micnicpm} or compressing larger models \citep{sheard_llama,llm-pruner}.
Some previous works \citep{li2023textbooks,bai2023qwen} emphasize the importance of prioritizing data quality over quantity when training smaller models from scratch. They demonstrate that small models can potentially outperform their larger counterparts with lower training costs. 
This insight offers a promising approach to training more powerful models with fewer resources.
From another perspective, model compression, which includes pruning, distillation, and quantization, are presented as a method to strike a balance between efficiency and performance for existing LLMs.
Pruning accelerates LLMs by removing non-essential parameters of the network with specialized hardware  \citep{llm-pruner,Frantar2023SparseGPTML,sun2024a}.
Distillation enables the model to acquire knowledge rapidly from a teacher model by mimicking the teacher's behavior \citep{Wu2023LaMiniLMAD,hsieh-etal-2023-distilling}.
Quantization can lower the costs of model storage and inference by converting the model to lower precision, more computationally efficient data types \citep{frantar2023optq,dettmers2024spqr}.

To accommodate low-resource and low-cost requirements, we introduce OpenBA-V2, an encoder-decoder Transformer model with 3.4B parameters.
OpenBA-V2 achieves a 77.3\% compression ratio of OpenBA \citep{li2023openba}, significantly lowering the resource requirements for deployment while maintaining high performance.
OpenBA-V2 adopts a multi-stage compression strategy that employs layer pruning or neural pruning at each stage, followed by a period of fast and efficient recovery training to minimize performance loss due to model compression.
After several compression stages, the model size has been reduced from 15B to 3.8B.
Subsequently, we use 700B tokens to continually pre-train the compressed model with an optimized objective, further boosting training efficiency and enhancing the model's capabilities.
Finally, we prune the model's vocabulary because of its redundancy, reducing the model size from 3.8B to 3.4B with almost no performance loss.
In addition, we have compiled a more extensive and diverse dataset from various sources compared to OpenBA.
These strategies have enabled OpenBA-V2 to achieve high performance with much fewer parameters.
Through OpenBA-V2, we aim to demonstrate that smaller models can achieve comparable performance to larger models through better training objectives and data strategies, facilitating the deployment across various industries of LLMs. Our code and model weights are available at \href{https://github.com/OpenNLG/OpenBA-v2}{https://github.com/OpenNLG/OpenBA-v2}.
\section{Related Work}
\subsection{Lightweight LLMs Pre-trained from Scratch}
Large language models (LLMs) have been proven to be very effective and have brought unprecedented success to various fields of artificial intelligence. 
Despite the trend towards developing larger models (with over a hundred billion parameters), lightweight LLMs have also played an essential role in today’s landscape for enabling efficient inference in limited hardware resources and edge devices.
Most lightweight LLMs were the by-product of the process when researchers explored the larger models 
and proposed alongside their larger version in a model family, such as OPT \citep{zhang2022opt}, BLOOMZ \citep{workshop2022bloom}, and more recent releases including GPT-Neo \citep{black2022gpt}, Galactica \citep{taylor2022galactica}, QWEN \citep{bai2023qwen}, as well as the LLM analyzing suite Pythia \citep{biderman2023pythia} and the transformer variant RWKV \citep{peng2023rwkv}. 
This type of lightweight LLMs generally follows the scaling law \citep{hoffmann2022empirical}, which recommends the relationship between model parameters and training data size for training LLMs. 
However, the recent evidence demonstrates that relatively small models, when trained with more data, can also match or even outperform their larger counterparts \citep{llama2}, indicating that when training smaller models for a longer time, the existing scaling law may not hold.
Therefore, researchers try to explore new lightweight LLMs with less than three billion parameters beyond the existing scaling law. 
These lightweight LLMs follow the core structure of popular LLMs but with their specific designs for more competitive performance such as adopting their own data-collecting strategy for a higher quality training corpus and extending the training phase for more tokens, including phi-1 \citep{gunasekar2023textbooks}, phi-1.5 \citep{li2023textbooks}, Falcon-RW \citep{penedo2023refinedweb}, Stable LM \citep{bellagente2024stable}, H2O-Danube \citep{singer2024h2o}, TinyLlama \citep{zhang2024tinyllama}, and even MobiLlama \citep{thawakar2024mobillama} with sub-billion parameters.
These lightweight models can reduce the inference budget and extend the applications of LLMs in real scenarios.





\subsection{Model Compression for LLMs}
Another way to achieve lightweight LLMs is model compression.
Model compression, including pruning, quantization, and distillation, achieves the trade-off between performance and efficiency based on existing outstanding LLMs.
Pruning reduces model parameters by eliminating modules, neurons, or individual connections that have a minimal impact on performance. 
Pruning can be categorized into unstructured and structured pruning. Unstructured pruning targets individual connections between neurons, resulting in sparse weight matrices that require specific hardware for efficiency gains \citep{Frantar2023SparseGPTML,sun2024a,ouderaa2024the}. 
Structured pruning removes entire rows or columns of weights, creating smaller dense matrices that are more hardware-friendly. However, it faces significant performance degradation at high compression ratios (> 30\%) \citep{ma2023llmpruner,zhang2023loraprune,an2023fluctuationbased}.
To improve the performance of the structured pruned model, \citet{sheard_llama} introduce the dynamic continual pre-training strategy, which resamples critical data for performance recovery for training.
Quantization uses lower-bit-width integers \citep{frantar2023optq,dettmers2024spqr,liu2024qllm,li2024loftq} or floats \citep{liu-etal-2023-llm,perez2023training} to represent weights, activations and KV caches.
It reduces the memory required for LLMs, which is essential when memory is a bottleneck for model deployment.
Currently, 8-bit quantization achieves nearly lossless performance compression and is compatible with most GPUs \citep{li2024evaluating}. However, researchers are not satisfied with this and continue to push the boundaries of ultra-low quantization precisions below 4-bit \citep{yuan2023rptq,ma2024era}.
In LLMs, many previous elaborate knowledge distillation strategies do not work \citep{jha2023train}, and much work now uses the data generated by large models directly as distillation signals for training smaller models and improving the small models' commonsense, reasoning ability, and so on \citep{Wu2023LaMiniLMAD,hsieh-etal-2023-distilling}.

\section{Dataset Preparation}

Compared to OpenBA~\citep{li2023openba}, we employ a more meticulous data processing process in OpenBA-V2 to ensure the quality of the training data.
Specifically, during the pre-training phase, we incorporate a greater diversity of pre-training data sources and combine more Chinese data, thereby further expanding the dataset domain distribution range. As for the instruction data, we introduce BiFlan-V2, which builds upon the BiFlan dataset~\citep{li2023openba} by implementing additional template designs and incorporating a wider variety of instructions.
We will introduce the collection and processing process of pre-training data and the instruction data in Sec.~\ref{sec:pretrain_data} and Sec.~\ref{sec:instruction_data}, respectively.

\subsection{Pre-training Data Collection and Processing}
\label{sec:pretrain_data}
\paragraph{Data Sources}
Due to the rapid development of the open-source community, there are quite a few publicly available pre-training data.
Considering the computation budget and data distribution, we collect a total of 4.4~TB pre-training data to enrich the data diversity and ensure comprehensive coverage.
Specifically, we collect English pre-training data from two sources: Pile \citep{gao2020pile} and RedPajama \citep{together2023redpajama}~\footnote{\url{https://huggingface.co/datasets/togethercomputer/RedPajama-Data-1T}}.
All the 22 diverse high-quality subsets of Pile are kept for pre-training, while we use six subsets of RedPajama: ArXiv, Books, C4, GitHub, StackExchange, and Wikipedia.
For Chinese pre-training data, we collect data from the following sources: an open-source version corpus released from Yuan \citep{wu2021yuan}, WanJuan \citep{he2023wanjuan}, SkyPile \citep{wei2023skywork}, CBook-150K~\footnote{\url{https://github.com/FudanNLPLAB/CBook-150K}}, Encyclopedias (i.e., Baidu Baike~\footnote{\url{https://baike.baidu.com/}}, Chinese Wikipedia~\footnote{\url{https://zh.wikipedia.org/wiki/}}) and Chinese Q\&A community~(Zhihu~\footnote{\url{https://www.zhihu.com/}}).
We use five subsets of WanJuan as pre-training data: ChinaNews-cn, Exam-cn, Law-cn, Patent-cn, and WebText-cn.

\paragraph{Data Processing}
Although we carefully select high-quality pre-training data sources and remove the same part from different sources, the rest may still have low-quality and repetitive data.
Therefore, we conduct the following data processing process to further improve the pre-training data quality and prevent potential risks:
\begin{itemize}
    \item \textbf{Privacy Filtering}: To prevent potential privacy leakage, we removed all phone numbers, email addresses, and web links from the collected pre-training data.
    \item \textbf{Deduplication}: Our pre-training data are collected from various open-sourced datasets. 
    To ensure data quality after merging, we employ deduplication strategies at multiple levels: document, character, and paragraph. 
    At the document level, each sample is treated as a document, and redundant documents are eliminated using a hash algorithm, thus retaining only unique documents. 
    Besides, at the paragraph level, we utilize a hash algorithm combined with an extra sentence segmenter to identify and remove duplicate sentences or paragraphs~(where consecutive 1-99 sentences are considered a paragraph). 
    Finally, at the character level, redundant characters are removed, and sequences of repeated characters are condensed to a single character.
    \item \textbf{Language Filtering} We utilize Polyglot~\footnote{\url{https://github.com/aboSamoor/polyglot}} to ascertain the language of the text, retaining only those texts confidently identified as either Chinese or English.
    This filtering process proves invaluable in filtering out gibberish, particularly for texts extracted from PDFs using OCR algorithms.
    \item \textbf{Internet Data Cleaning} The data collected from the Internet frequently contains incompletions, unrecognizable characters, and web page tags. 
    Consequently, we implement filtering procedures to remove sentences containing fewer than 10 words and filter out unusual characters and HTML tags.
\end{itemize}

\begin{figure}[t]
	\centering
	\subfigure[]{
		\begin{minipage}[t]{0.5\linewidth}
			\centering
			\includegraphics[width=2.3in]{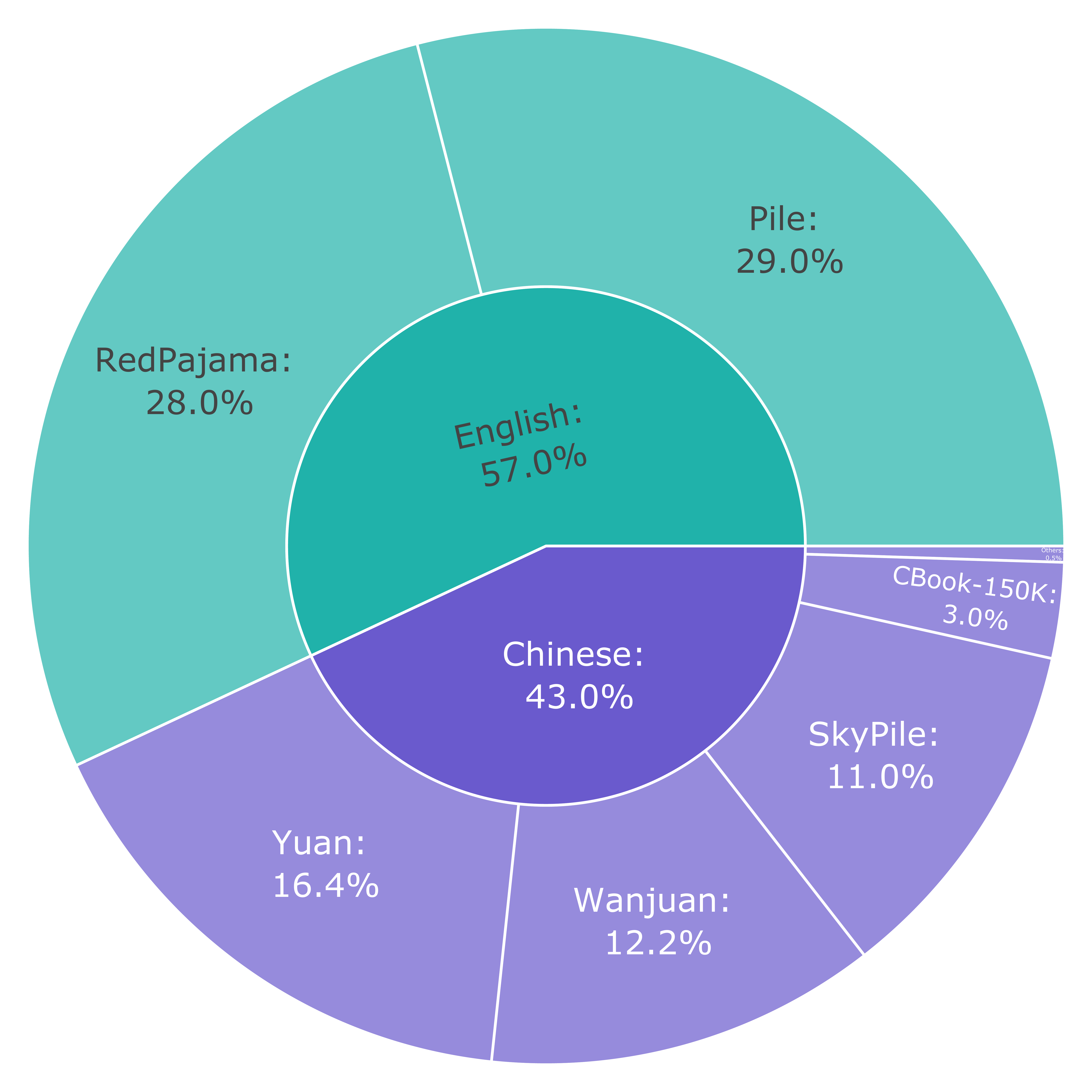}
		\end{minipage}
		\label{fig:pretrain_data}
	}%
	\subfigure[]{
		\begin{minipage}[t]{0.5\linewidth}
			\centering
			\includegraphics[width=2.3in]{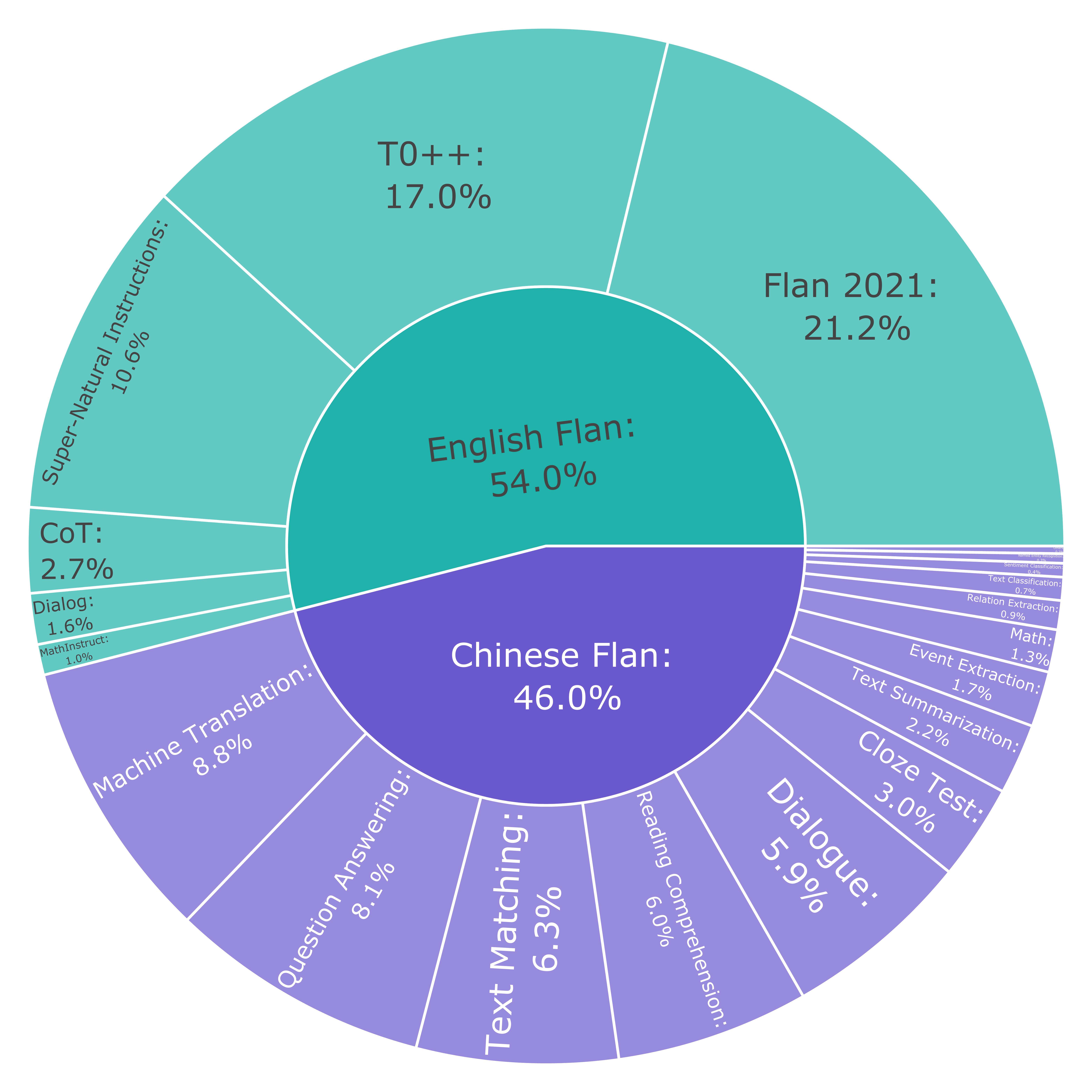}
		\end{minipage}
		\label{fig:instruction_data}
	}%
	\centering
	\caption{Data distribution of different training corpus, where Figure (a) shows the distribution of the pre-training data and Figure (b) shows the distribution of the BiFlan-V2 instruction data.}
	\label{fig:data_distribution}
\end{figure}

\paragraph{Data Statistics}
All the pre-training data mentioned above requires 4.4~TB disk space to save, and the final pre-training data consists of 57.0\% English data and 43.0\% Chinese data. The pre-training data distribution is illustrated in Fig.~\ref{fig:pretrain_data}.

\subsection{BiFlan-V2: Instruction Data Collection}
\label{sec:instruction_data}

\paragraph{English Instruction Data Collection}
Following the distribution and collection source of the BiFlan dataset introduced in OpenBA~\citep{li2023openba}, our English instruction data is mainly collected from the Flan Collection \citep{chung2022scaling,longpre2023flan}.
The Flan Collection encompasses more than 1800 tasks, which is currently the most comprehensive instruction collection.
We follow the official guidelines to collect and process the English Flan collection with two steps,
i.e., downloading five sub-mixtures from the Flan Collection and then combining them according to the
specified mixture rates~\footnote{\url{https://github.com/google-research/FLAN/tree/main/flan/v2}}.
Besides, we also incorporate the MathInstruct dataset~\citep{yue2023mammoth} ~\footnote{\url{https://huggingface.co/datasets/TIGER-Lab/MathInstruct}} to improve the model reasoning ability.

\paragraph{Chinese Instruction Data Collection}
For the Chinese instruction data, apart from the Chinese Flan data introduced in OpenBA sourcing from various competitions, academic papers, and open-source projects, we incorporate more math reasoning, text-matching, question-answering, reading comprehension, and event-extraction data in this version.
Besides, we also use BELLE School Math dataset \citep{BELLE,belle2023exploring,wen2023chathome} ~\footnote{\url{https://huggingface.co/datasets/BelleGroup/school_math_0.25M}} to improve the math reasoning abilities.
Finally, the Chinese Instruction data is collected from 44 different Chinese tasks with 50 million data entries. 
The Chinese instructions for each task are still designed manually.

\paragraph{BiFlan-V2 Data Statistics}
We show the instruction data distribution in Fig.~\ref{fig:instruction_data}. Following OpenBA, we filter out samples with lengths exceeding the encoder’s maximum length, ensuring the critical parts of instructions are not truncated. Finally, the instruction data consists of 54.0\% English data and 46.0\% Chinese data.
\section{Fast Multi-Stage Pruning}
\label{sec:msp}

In this section, we introduce the Fast Multi-Stage Pruning, including layer pruning, neural pruning, vocabulary pruning, and the objectives used in each stage.

\subsection{Pruning Strategies}
\subsubsection{Layer Pruning}
Layerdrop \citep{layerdrop,layerdrop2} is a straightforward method for pruning Transformer model parameters by randomly dropping entire layers of the model.
Layer pruning does not severely damage the model's architecture. 
Therefore, a pruned model can maintain a relatively low perplexity (PPL) even without recovery training. Thus, we choose to perform layer pruning first. We conduct preliminary experiments to determine the optimal layers to prune, leading to the following insights:
\begin{itemize}
    \item Compared with the top and bottom layers, pruning intermediate layers causes less damage to the model, which is also shown in LLM-Pruner \citep{llm-pruner}.
    \item Pruning layers with more intervals cause less damage to the model.
\end{itemize}
Additionally, we investigate the effects of the quantity of pruned parameters on the model's performance. We observe that once the quantity of pruned parameters reaches a certain threshold, the model's performance will plummet.
Consequently, we adopt a staged approach to pruning the model's layers. After each stage, we use a small amount of data to recover the model performance.

\begin{table}\small
\centering
\label{tab:stages_overview}

\begin{tabular}{l|c|cccc}
  \toprule
  \textbf{Models} & \textbf{\#Params}  & \textbf{Pruned-Enc} & \textbf{Pruned-Dec} & \bf Loss After Prune & \bf Loss After training \\
   \midrule
  OpenBA & 15B & - & - &  - & 1.73 \\
  \midrule
  Direct Prune & 9.9B & all & all & 2.69 & - \\
\midrule
  Stage1 & 12.3B & [4,8] & [7,11,18,20,22,30] & 1.85 & 1.76 \\

  Stage2 & 11.0B & [6,10] & [4,16,27] &  1.90 & 1.80 \\
    
  Stage3 & 9.9B & - & [10,24,33] &  1.89 & 1.82  \\
  \bottomrule
\end{tabular}
\caption{Overview of the layer pruning process.}
\label{tab:layer_pruning_process}
\end{table}

\subsubsection{Neural Pruning}
Existing works have explored how to prune the matrix parameters of the model while minimizing performance degradation \citep{llm-pruner,wanda,magnitude,sheard_llama}. 
Some of the works attempt to compute the importance of each element in the parameter matrix and zero out some of these elements, utilizing coefficient matrices for computation.

The importance of parameters often correlates with their absolute values, parameter gradients, and activation values. Therefore, these methods typically require the model to undergo forward and backward propagation on a certain amount of data.
Another limitation is that these methods do not truly prune the parameter matrix to another shape; instead, they zero out some parameters and utilize sparse matrix operations to perform model inference.
Such methods have minimal impact on the model when pruning a small number of parameters. However, when the target pruning amount reaches 30\% or more, the model's performance will drop sharply. Moreover, since they only sparsify the matrix and do not facilitate subsequent retraining, this becomes disadvantageous.

\begin{figure}[t]
    \centering
    \includegraphics[width=0.98\hsize]{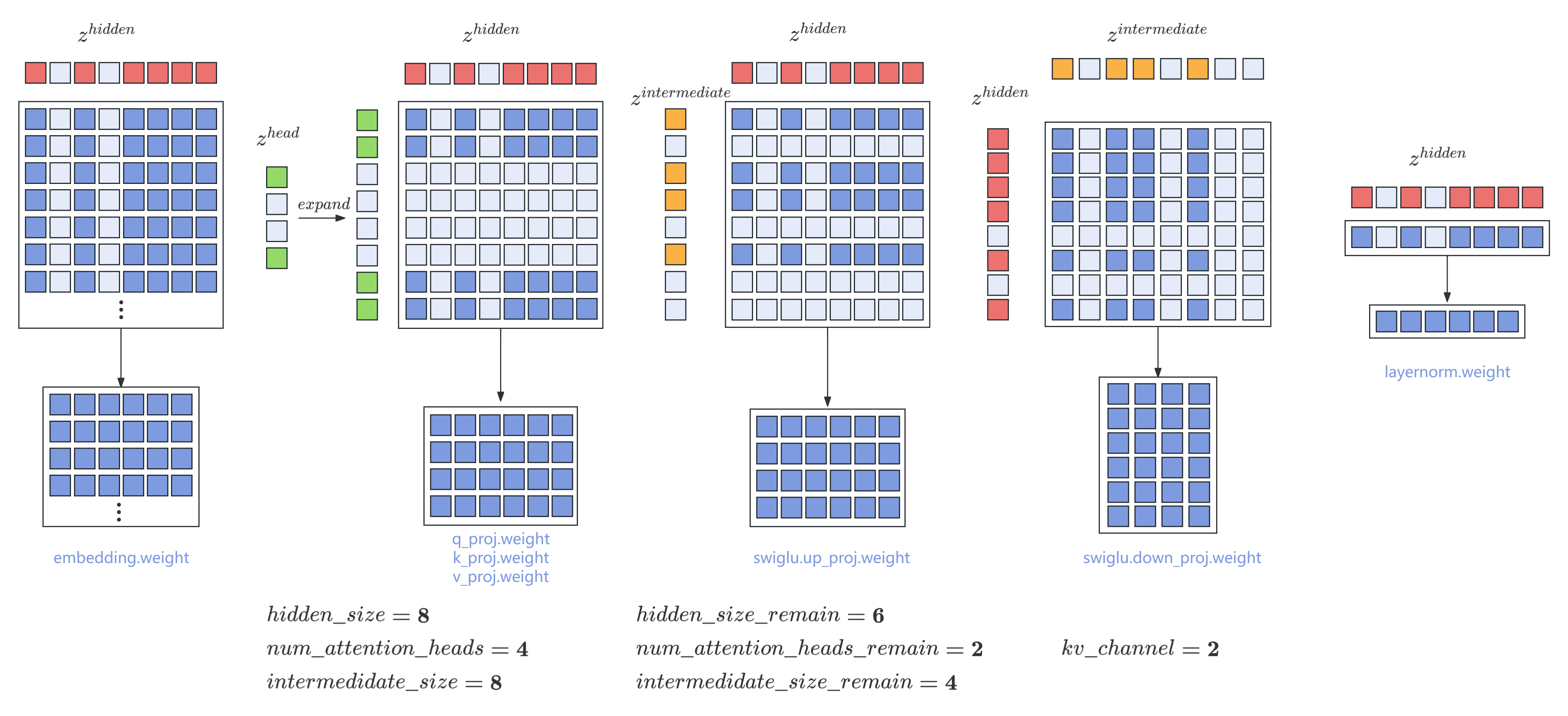}
    \caption{Illustraion of random neural pruning.}
    \label{fig:neural_prune}
\end{figure}

Both Sheard-LLaMA \citep{sheard_llama} and LLM-Pruner \citep{llm-pruner} reveal that when pruning the parameter matrix of the model into another shape of a dense matrix, we should not disrupt the dependent structures within the model. The dependent structures have been clearly defined in \citet{llm-pruner}. 
Furthermore, experiments from various studies have shown that the method of pruning, whether following specific importance criteria or being random, has little impact on the model's performance after it has been pruned and subsequently retrained for a while.
Therefore, our approach involves directly randomly pruning the rows and columns of the matrix based on dependent structures and the dimensions of the target model. Figure \ref{fig:neural_prune} can help better understand our method.

\subsubsection{Vocabulary Pruning}
The original OpenBA model employs a multilingual vocabulary comprising approximately 260,000 tokens. However, it is primarily trained on Chinese and English corpora and is designed to serve the Chinese-English language pair exclusively. Consequently, many tokens in the vocabulary exhibit very low usage frequencies, resulting in a considerable number of idle or rarely used embedding vectors within the model's embedding matrix.

\begin{figure}[t]
    \centering
    \includegraphics[width=0.7\hsize]{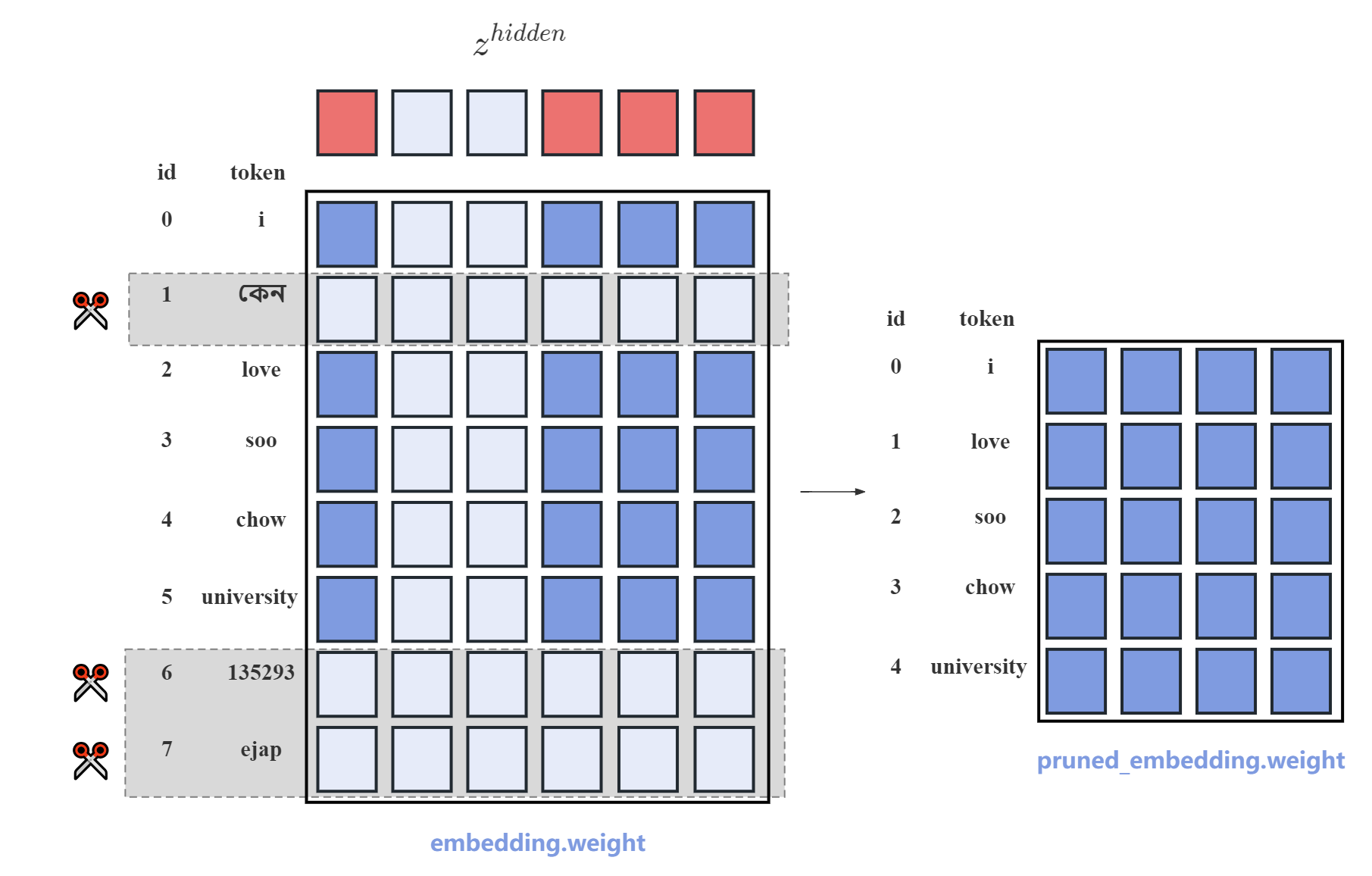}
    \caption{Illustraion of how to prune the vocabulary and the corresponding embedding.}
    \label{fig:emb_prune}
\end{figure}

To address this issue, we conduct a comprehensive analysis of token occurrences in the pre-training corpus and sort all tokens based on their frequency of occurrence. Then, we retain the top K tokens with the highest occurrence frequencies while pruning the remaining tokens. Additionally, we prune the embedding associated with these tokens and reorganized the token IDs and embedding matrices accordingly. This approach enables us to further reduce the number of parameters in the model, thereby decreasing the memory footprint.
Figure \ref{fig:emb_prune} shows how we prune the embedding weights and rearrange the token IDs according to the pruned embedding.
\subsection{Training Objective}

\subsubsection{UL2}
The 15B OpenBA model employs the UL2 training strategy, a mixture of denoisers approach proposed by \citep{UL2}, which requires the model to reconstruct sentences in various types of noise.
\begin{itemize}
\item \textbf{R-Denoising} Regular denoising is the standard span corruption that sets a range of 2 to 5 tokens as the masked span length and masks ratio about 15\% of the input tokens. This denoising task is relatively simple since the span is short and efficient for the model to acquire knowledge embedded in the text.
\item \textbf{S-Denoising} Sequence denoising aims to endow the model with generation capability, where the input text is split into two sub-sequences, and the model should predict the latter sequence conditioned on the first sequence. In the S-Denoising setting, the model can acquire the generation ability.
\item \textbf{X-Denoising} To bridge the gap between the R-Denoising and S-Denoising, X-Denoising can be viewed as an extreme version of denoising, where approximately 50\% of the input sequence is masked by increasing either the masked span length or the corruption rate. 
Such a denoising strategy simulates the situation where a model needs to generate long targets from memory with relatively limited information.
\end{itemize}
The detailed information for corruption ratio and span length can be found in Table~\ref{tab:ul2_denoiser_setting}.
\begin{table}[ht]
    \centering
    \small
    \begin{tabular}{l c c c c}
    \toprule
    \bf Type & \bf Span Length ($\mu$) & \bf Corruption Ratio (\%) & \bf \#Num & \bf Sentinel \\
    \midrule
     <R>-1 &  3 & 15.0 & $K$ & \texttt{<R>}\\
     <R>-2 &  8 & 15.0 & $K$ & \texttt{<R>}\\
     <S> & - & 25.0 & 1 & \texttt{<S>} \\
     <X>-1 & 3  & 50.0  & $K$ & \texttt{<X>} \\
     <X>-2 & 8  & 50.0  & $K$ & \texttt{<X>} \\
     <X>-3 & 64 &  15.0 & $K$ & \texttt{<X>} \\
     <X>-4 & 64 & 50.0  & $K$ & \texttt{<X>} \\
    \bottomrule
    \end{tabular}
    \caption{Details of different noise type in UL2 objective.}
    \label{tab:ul2_denoiser_setting}
\end{table}

\subsubsection{Dynamic-UL2}

\begin{algorithm}[ht]
\setlength{\baselineskip}{13bp}
\caption{Dynamic-UL2}\label{alg:two}
\SetKw{Require}{Require}  
\SetKwProg{Subroutine}{Subroutine}{}{}
\SetKwFunction{Update}{UpdateWight}   
\Require: Training dataset $D$, validation data $D_1^{\text {val}}, D_2^{\text {val}}, \cdots, D_7^{\text {val }}$, where $D_{i}^{\text {val }}$ denote the valid data with noise type $i$, the initial $UL2$ noise prop $p_0 \in \mathbb{R}^7$, the $\ell_{\text {ref }} \in \mathbb{R}^7$, LM loss function $\mathcal{L}$, noise adding function $\mathcal{F}, $training steps $T$, evaluation interval $m$, model parameters $\theta$.\\

\For{$t=1, \cdots, T$}{
\If {$t \bmod m=0$}{
$\ell_t[i] \leftarrow  \mathcal{L}\left(\theta, D_i^{\text {val }}\right)$ \Comment{Calculate the loss of each noise type}\\

$\Delta_t[i] \leftarrow \max \left\{\ell_t[i]-\ell_{\text {ref }}[i], 0\right\} \quad $ \Comment{Calculate loss difference} \\
$p_t \leftarrow \Update \left(p_{t-m}, \Delta_t\right) \quad $ \\
}
Sample a batch of data $B$ from $D$ \\
Adding noise to $B$ to obtain $\hat{B}$ according to $p_t$ \\

Update $\theta$ with $\mathcal{L}(\theta, \hat{B}$)\\
}
\Subroutine{\Update($p,\Delta$)}{
$\alpha \leftarrow p \cdot \exp (\Delta)$ \\
$p \leftarrow \frac{\alpha}{\sum_i \alpha[i]}$ \Comment{Calculate the new noise ratio for each noise type} \\
\KwRet{$p$}  \\
}  
\KwRet{$\theta$}

\end{algorithm}

Previous studies have shown that model pruning can affect different capabilities to different extents. Consequently, works such as \citet{sheard_llama,doremi} suggest dynamically adjusting the sampling ratio for each domain according to the model's loss.

The UL2 objective utilized in OpenBA incorporates various types of noise, with each denoising process specifically training the model to enhance specific capabilities. For example, the <S> task improves the model's capabilities in text continuation and generation, while the <S> and <X> tasks help the model attain better comprehension and extraction capabilities.
Inspired by previous works, we propose Dynamic-UL2, which dynamically adjusts the proportion of each type of noise based on the loss for each noise on the valid set.
The Dynamic-UL2 Algorithm can be found in Algorithm \ref{alg:two}.

\subsubsection{Optimized-UL2}
\begin{figure}[t]
    \centering
    \includegraphics[width=0.98\hsize]{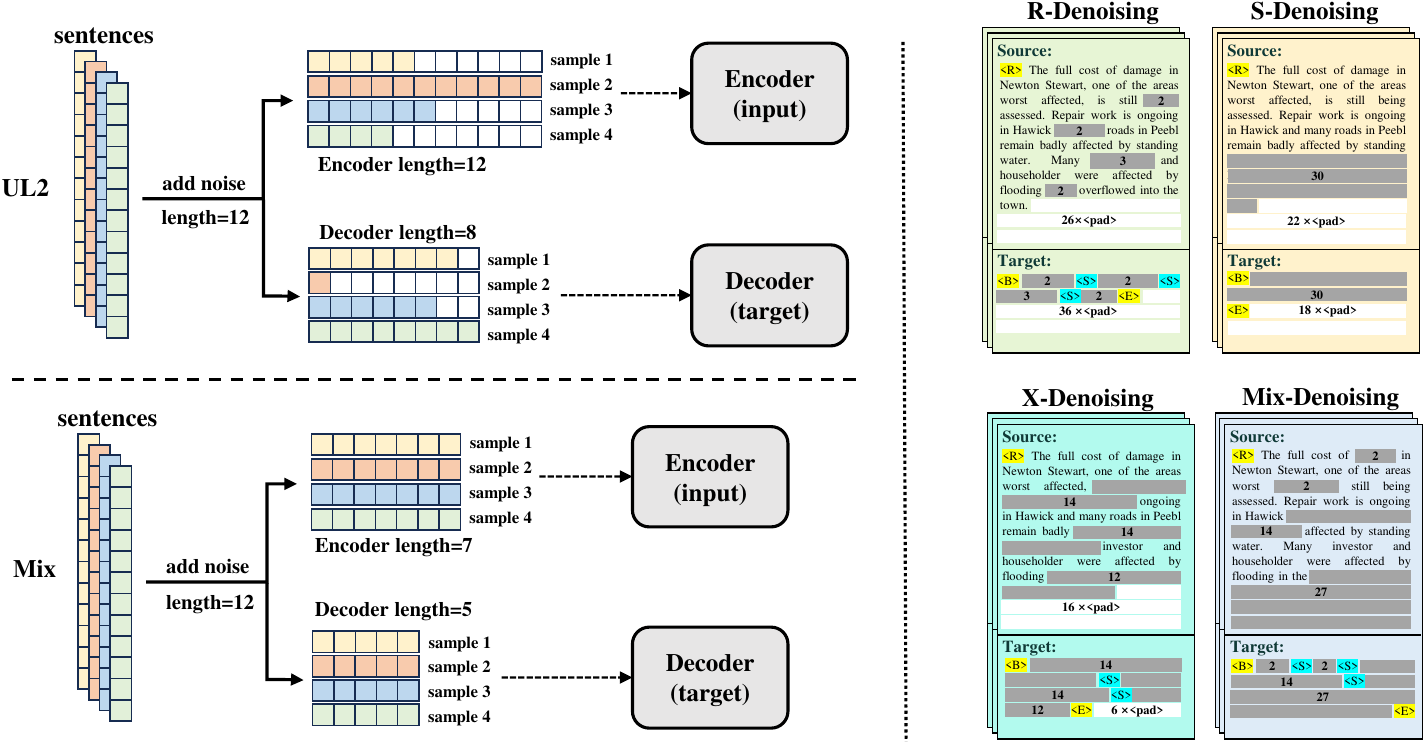}
    \caption{Illustration of Optimized-UL2, which allows training with various types of noise with few padding tokens.}
    \label{fig:O-UL2}
\end{figure}
While UL2 demonstrates significant performance enhancements by integrating various noise types, mixing multiple noises necessitates extensive padding tokens to accommodate diverse noise types within a batch.
Therefore, the training efficiency of UL2 is relatively low.

As shown in Figure 1, different noise types will influence the input length of the encoder and decoder.
For a given sentence, if more tokens need to be masked based on the selected noise type, the length of the encoder will increase while the length of the decoder will decrease, and vice versa.
Hence, we must pad the input of the encoder and decoder to a predefined maximum length throughout the training process.
Our estimations show that roughly 40\% of the tokens in UL2 are padding tokens, impeding actual training efficiency.

To address this issue, we keep a global mask rate and fuse various noise types into a single sentence called Mix-Denoising.
Specifically, we initially apply S-noise to introduce noise to a single sentence. Subsequently, we employ R-noise and X-noise to add noise to the rest of the sentence, ensuring the total number of masked tokens reaches a predefined target.
For S-noise, we set a lower and an upper bound, and randomly sample the number of masked tokens from a normal distribution within these limits.
For R-noise and X-noise, we can calculate the number of tokens that need to be masked based on the length of the rest of the sentence.
For simplicity, we select [X] as the prefix token.
With Mix-Denoising, we can keep the input length of the encoder and decoder as a fixed value through the whole training process with few padding tokens, thus improving the training efficiency of UL2.
However, Mix-Denoising may cause the model to lose some generation ability, as R-noise and X-noise have disrupted the natural language order of its decoder.
Therefore, we preserve the original S-Denoising task to enhance the model's generation capability.
It is worth noting that we fix the number of masked tokens for S-noise to ensure consistency.
Furthermore, Mix-Denoising has more sentinel tokens than S-Denoising, resulting in more input tokens. To standardize the input lengths between the two tasks, we truncate the original sentence in Mix-Denoising. During training, Mix-Denoising and S-Denoising comprise about 20\% and 80\% of our training data, respectively.

These adaptations mitigate the need for excessive padding, achieving model performance comparable to UL2 with almost no padding token.
Our approach has more valid tokens than UL2, thus enabling superior performance at equivalent training costs.

\section{Implementation}
\subsection{Model Architecture}
Despite the reduction in model parameters, OpenBA-V2 maintains the same model architecture as OpenBA \citep{li2023openba}, including an Encoder-Decoder model structure, rotary embedding scheme \citep{su2024roformer},and the SwiGLU Activation Function \citep{shazeer2020glu}.

\subsection{Training}
\begin{table}\small
\centering
\label{tab:stages_overview}
\begin{tabular}{l|c|ccccc}
  \toprule
  \textbf{Models} & \textbf{\#Params}  & \textbf{Enc} & \textbf{Dec} & \textbf{Hidden} & \textbf{FFN} & \textbf{Heads}  \\
   \midrule
  OpenBA & 15B & 12 & 36 & 4,096 & 11,008 & 40 \\
\midrule
  Stage1 & 12.3B & 10 & 30 &  4,096 & 11,008 & 40  \\

  Stage2 & 11.0B & 8 & 27 &  4,096 & 11,008 & 40  \\

  Stage3 & 9.9B & 8 & 24 &  4,096 & 11,008 & 40  \\
  Stage4 & 3.8B & 8 & 24 &  2,560 & 6,912 & 20  \\
  \midrule 
  Stage5 & 3.8B & 8 & 24 &  2,560 & 6,912 & 20  \\
 
   \toprule
   \textbf{Models} & \textbf{\#Params}  & \textbf{Enc-Len} & \textbf{Dec-Len} & \textbf{Tokens} & \textbf{Objective} & \textbf{Flops} ($\times 10^{20}$)  \\
    \midrule
     OpenBA & 15B & 570 & 380 &  350B & UL2 & 277.1  \\
\midrule
   Stage1 & 12.3B & 570 & 380 &  10B & D-UL2 & 6.7  \\

  Stage2 & 11.0B & 570 & 380 &  10B & D-UL2 & 5.9  \\

  Stage3 & 9.9B & 570 & 380 &  15B & D-UL2 & 8.1  \\
  Stage4 & 3.8B & 570 & 380 &  65B & D-UL2 & 13.0  \\
  \midrule 
  Stage5 & 3.8B & 1024 & 1024 &  700B & O-UL2 & 99.1 \\
  \bottomrule
\end{tabular}
\label{tab:training_stages}
\caption{Overview of our entire compression and training process, including the model architecture at each stage, the objective functions, the scale of the training data, and computational requirements.}
\end{table}
We first use a relatively small amount of tokens to compress the 15B model to 3.8B without a significant loss in model capability. Subsequently, we use a large number of tokens to train the model further for better performance. The entire process can be divided into multiple stages, and table \ref{tab:training_stages} illustrates the model sizes and training objectives at different stages.
We use a cosine scheduler for stages 1-4 with the max learning rate 1e-4 and the min learning rate 5e-5.
For stage 5, we use the max learning rate 5e-5 and the min learning rate 1e-5.
After pruning, We directly prune 140,000 tokens from the vocabulary, reducing the model size from 3.8B to 3.4B.
\section{Experiments \& Evaluation}
\subsection{Baseline Models}
We mainly select open-source models of approximately 3B parameters for comparing model performance. Additionally, we include some state-of-the-art (SOTA) models with around 7B parameters.
The details of the selected models are shown in table \ref{tab:baseline_models}.
\begin{table}[ht]
    \centering
    \small
    \begin{tabular}{l c c c c}
    \toprule
    \bf Model & \bf \#Param. & \bf \#Tokens & \bf Open-Sourced Data & \bf Language \\
    \midrule
    LLaMA2 \cite{llama2} & 7B & 2T & NO & EN* \\
    Baichuan2 \cite{baichuan2} & 7B & 2.6T & NO & ZH,EN \\
    ChatGLM3 \cite{glm} & 6B & - & NO & ZH,EN \\
    Chinese-LLaMA2 \cite{chinese-llama2} & 7B & - & NO & ZH,EN \\
    TinyLlama \cite{zhang2024tinyllama}&  1.1B & 3T & YES & EN \\
    OpenLLaMA-v2 \cite{openllama} & 3B & 1T & YES & EN \\
    BLOOM \cite{bloom} & 3B & 350B & YES  & Multi-Lingual \\
    MindLLM \cite{mindllm} & 1.3B & 323B & YES & ZH,EN \\ 
    GPT-NEO \cite{GPT-NEO} & 2.7B & 420B & YES & EN \\
    INCITE-Base \cite{redpajama} & 3B & 800B & YES & EN\\
    Sheard-LLaMA \cite{sheard_llama} & 2.7B & 50B & YES & EN \\
    Phi2 \cite{li2023textbooks} & 2.7B & 1.4T & NO & ZH,EN \\
    Qwen \cite{bai2023qwen} & 1.8B & 2.2T &  NO & ZH,EN \\
    Mini-CPM \cite{micnicpm} & 2.7B & 1.1T & NO & ZH,EN \\
    \bottomrule
    \end{tabular}
    \caption{Details of the baseline models. We denote the training corpus of LLaMA2 with * because English significantly predominates in terms of proportion among all languages.}
    \label{tab:baseline_models}
\end{table}

\subsection{Settings}
We select a diverse range of tasks for evaluation. For world common knowledge, we evaluate the models on MMLU \citep{MMLU}, CMMLU \citep{cmmlu}, C-EVAL \citep{c-eval} and BBH \citep{bbh}. For commonsense reasoning and reading comprehension, we select SciQ \cite{sciq}, PIQA \cite{piqa}, ARC \cite{arc}, LogiQA \cite{logiqa}, and BoolQ \cite{BoolQ}. We annotate the num-shots used during evaluation in the parentheses on the right of the dataset name. For the performance of the baseline models, if results for the corresponding dataset are available in the original paper, we directly report those results; otherwise, we report the results we reproduced. We use the LM-Evaluation-Harness repository \cite{lm-eval} to reproduce the results of the baseline models.
Since LM-Evaluation-Harness is not well adapted for encoder-decoder architecture, we have modified the original repository and redeveloped the evaluation code for OpenBA-V2.


\begin{table}[ht]
    \centering
    \small
    \resizebox{\textwidth}{!}{
    \begin{tabular}{l c c c c c c c}
    \toprule
    \multirow{2}{*}{\bf Model} & \multirow{2}{*}{\bf \#Param.} & \multicolumn{3}{c}{\bf English Benchmark} & \multicolumn{3}{c}{\bf Chinese Benchmark} \\
    \cmidrule(r){3-5} \cmidrule(r){6-8}
    & & \bf Avg. (EN) & \bf MMLU(5) & \bf BBH(3) & \bf Avg. (ZH) & \bf C-EVAL(5) & \bf CMMLU(5) \\
    \midrule
    LLaMA2 & 7B & 38.8 & 44.3 & 33.2 & - & - & - \\
    Chinese-LLaMA2 & 7B & - & - & - & 34.5 & 35.9 & 33.0 \\
    Baichuan2 & 7B & 47.9 & 54.2 & 41.6 & 55.6 & 54.0 & 57.1\\ 
    ChatGLM3 & 6B & 63.8 & 61.4 & 66.1 & 68.3 & 69.0 & 67.5 \\ 
    \midrule
    TinyLlama & 1.1B & 26.1 & 25.4 & 26.8 & - & - & - \\ 
    OpenLLaMA-v2 & 3B & 27.9 & 26.2 & 29.5 & - & - & - \\
    BLOOM & 3B & 23.2 & 25.8 & 20.5 & 25.3 & 25.4 & 25.2 \\ 
    MindLLM & 1.3B & 17.7 & 25.4 & 9.9 & 26.9 & 28.0 & 25.7 \\ 
    GPT-NEO & 2.7B & 25.9 & 25.0 & 26.7 & - & - & - \\
    INCITE-Base & 3B & 27.0 & 27.0 & 27.0 & - & - & - \\
    Sheard-LLaMA & 2.7B & 28.2 & 27.1 & 29.2 & - & - & - \\
    \midrule
    Qwen & 1.8B & 33.8 & 45.3 & 22.3 & 54.1 & 56.1 & 52.1 \\
    MiniCPM & 2.7B & 45.2 & 53.5 & 36.9 & 51.1 & 51.1 & 51.1 \\
    Phi2 & 2.7B & 56.4 & 56.9 & 59.1 & 33.8 & 35.1 & 32.5 \\ 
    \midrule
    OpenBA & 15B & 37.2 & 40.2 & 34.1 & 40.7 & 39.8 & 41.5 \\
    OpenBA-V2 & 3.8B & 34.2 & 38.4 & 29.9 & 38.4 & 38.3 & 38.5 \\
    OpenBA-V2$\dag$ & 3.4B & 34.2 & 38.4 & 30.0 & 38.4 & 38.3 & 38.5\\
    \bottomrule
    \end{tabular}}
    \caption{Model performance on world knowledge tasks with different languages (English and Chinese), where $\dag$ denotes the model with pruned vocabulary (the vocab size is 12k). }
    \label{tab:main_res1}
\end{table}


\begin{table}[t]
    \centering
    \small    
    \setlength\tabcolsep{3pt}
    \begin{tabular}{l c c c c c c c c }
    \toprule
    \bf Model & \bf \#Param. & \bf AVG. & \bf SciQ(0) & \bf PIQA(0) & \bf ARC-E(0) & \bf ARC-C(25) & \bf LogiQA(0) & \bf BoolQ(32) \\
    \midrule
    LLaMA2 & 7B & 69.0 & 93.7 & 78.1 & 76.4 & 53.0 & 30.7 & 82.1 \\ 
    Baichuan2 & 7B & 67.4 & 94.7 & 78.4 & 78.9 & 49.7 & 29.3 & 73.2 \\ 
    ChatGLM3 & 6B & 66.3 & 94.4 & 73.5 & 68.6 & 52.3 & 30.3 & 78.7 \\ 
    \midrule
    TinyLlama & 1.1B & 56.1 & 94.0 & 73.3 & 55.3  & 30.1 & 26.3 & 57.8 \\ 
    OpenLLaMA-v2 & 3B & 61.9 & 91.8 & 76.2 & 66.5 & 39.0 & 28.1 & 69.6 \\ 
    BLOOM & 3B & 59.2 & 93.5 & 70.7 & 64.2 & 35.2 & 29.3 & 62.1 \\ 
    MindLLM & 1.3B & 49.5 & 80.2 & 64.9 & 47.1 & 24.8 & 28.0 & 52.1 \\ 
    GPT-NEO & 2.7B & 59.7 & 93.3 & 74.2 &  65.3 & 35.2 & 28.4 & 61.8 \\
    INCITE-Base-3B & 3B & 61.1 & 90.7 & 74.6 & 67.7 & 40.2 & 27.7 & 65.9 \\
    Sheard-LLaMA & 2.7B & 62.9 & 90.8 & 75.8 & 67.0 & 41.2 & 28.9 & 73.7 \\
    \midrule
    Qwen & 1.8B & 61.1 & 92.2 & 73.6 & 63.7 & 38.5 & 31.6 & 66.8 \\
    MiniCPM & 2.7B & 71.8 & 96.0 & 76.6 & 85.4 & 68.0 & 31.2 & 73.7  \\
    Phi2 & 2B & 73.5 & 97.1 & 79.3 & 85.2 & 61.2 & 33.8 & 84.2  \\
    \midrule
    OpenBA & 15B & 68.7 & 94.6 & 72.0 & 69.7 & 60.0 & 33.3 & 82.4 \\
    OpenBA-V2 & 3.8B & 63.4 & 94.7& 70.0 & 63.9	& 45.4 & 31.6 & 74.6 \\
    OpenBA-V2 $\dag$ & 3.4B & 63.4 & 94.6 & 70.4 & 63.8	& 45.4 & 31.6 & 74.7 \\
    \bottomrule
    \end{tabular}
    \caption{Model performace on commonsense \& reading comprehension tasks. $\dag$ denotes the model with pruned vocabulary. }
    \label{tab:main_res3}
\end{table}

\subsection{Main Results}
We compare the model performance with other baselines in Table~\ref{tab:main_res1} and~\ref{tab:main_res3}.
In each table, the first group includes models with more than 6B parameters. The second group includes models with 3B or fewer parameters trained on open-sourced data. The third group includes models with 3B or fewer parameters trained on Non-open-sourced data. The final group includes all versions of OpenBA models.

Overall, OpenBA-V2 outperforms all models with 3B or fewer parameters trained on open-sourced data, demonstrating its strong competitiveness among models of the same size.
OpenBA-V2 is weaker than larger models, but it still demonstrates its competitiveness.
For example, OpenBA-V2 surpasses Chinese-LLaMA2 on the Chinese benchmark and shows no significant gap compared to other larger models in commonsense reasoning and reading comprehension tasks.
Compared to models with fewer than 3B parameters trained on non-open-source data, OpenBA-V2 outperforms them only in specific tasks, and there is a gap in most scenarios, highlighting the importance of high-quality data for LLMs.
Compared to OpenBA, OpenBA-V2 achieves strong performance with just 23\% parameters, indicating a significant potential for model compression.
The model can be more lightweight and cost-effective by implementing effective compression strategies and recovery training.

\subsection{NER Finetuning Performance}

We further explore the potential of the OpenBA model series for Named-Entity-Recognition (NER).
We utilize the Pile-NER \citep{zhou2023universalner} dataset as our training dataset, comprising approximately 240,000 entities across 13,000 distinct categories. 
Following this, we evaluate the model on the MIT \citep{liu2013asgard} and CrossNER \citep{liu2021crossner} datasets, where the entity labels are mostly unseen during the training phase.
We adopt the method outlined in GNER \citep{ding2024rethinking} for incorporating negative instances into the training phase.
Additionally, we compare our approach with three baselines: sheard-llama \citep{sheard_llama}, open-llama-v2 \citep{openllama} and OpenBA-15B \citep{li2023openba}, employing identical training methodologies and strict entity-level $F_1$ score as evaluation metrics.
The results are summarized in table~\ref{tab:ner_results}.
Our model significantly outperforms the others, achieving superior performance that exceeds the original 15B model before its pruning.

\begin{table}[ht]
    \small
    \centering
    \begin{tabular}{l | c c c c c c c | c}
    \toprule
    \bf Model & \bf Movie & \bf Rest. & \bf AI & \bf Literature & \bf Music & \bf Politics & \bf Science & \bf Avg. \\
    \midrule
    Sheard-LLaMA & 29.4 & 15.7 & 34.5 & 33.7 & 39.7 & 33.3 & 35.8 & 31.7 \\
    OpenLLaMA-v2 & 63.0 & 44.3 & 58.9 & 50.2 & 72.0 & 66.4 & 62.5 & 59.6 \\
    \midrule
    OpenBA-NER-15B & 51.4 & 37.4 & 58.5 & 54.6 & 69.1 & 68.4 & 68.2 & 58.2 \\
    OpenBA-V2-NER-3B & 60.3 & 43.3 & 63.8 & 54.3 & 72.4 & 67.4 & 70.6 & 61.7 \\
    \bottomrule
    \end{tabular}
    \caption{Zero-shot performance in out-of-domain evaluation on MIT \citep{liu2013asgard} and CrossNER \citep{liu2021crossner} datasets.}
    \label{tab:ner_results}
\end{table}

\section{Analysis}
\label{sec:ana}
In this section, we mainly illustrate the motivation behind multi-stage pruning~(Sec.~\ref{motivation_multi_stage_pruning}), the effectiveness of Dynamic-UL2 strategy~(Sec.~\ref{effectiveness_of_dul2}), as well as the impact of vocabulary pruning on the final performance~(Sec.~\ref{subsec:vocab_pruning}).

\subsection{Motivation behind Multi-Stage Pruning}
\label{motivation_multi_stage_pruning}
\begin{figure}[t]
	\centering
	\subfigure[Layer Pruning]{
		\begin{minipage}[t]{0.5\linewidth}
			\centering
			\includegraphics[width=2.3in]{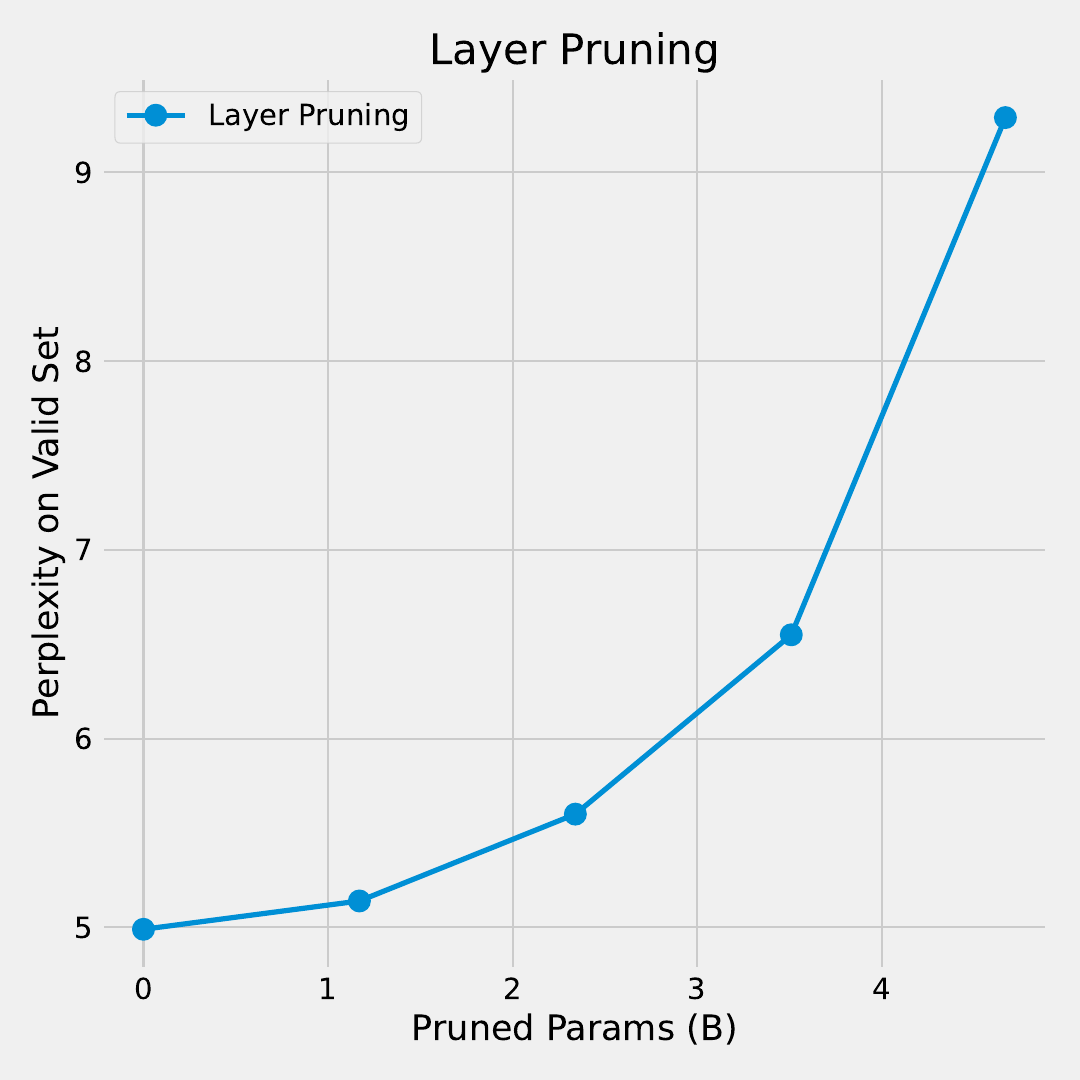}
		\end{minipage}
		\label{fig:2a}
	}%
	\subfigure[Neural Pruning]{
		\begin{minipage}[t]{0.5\linewidth}
			\centering
			\includegraphics[width=2.3in]{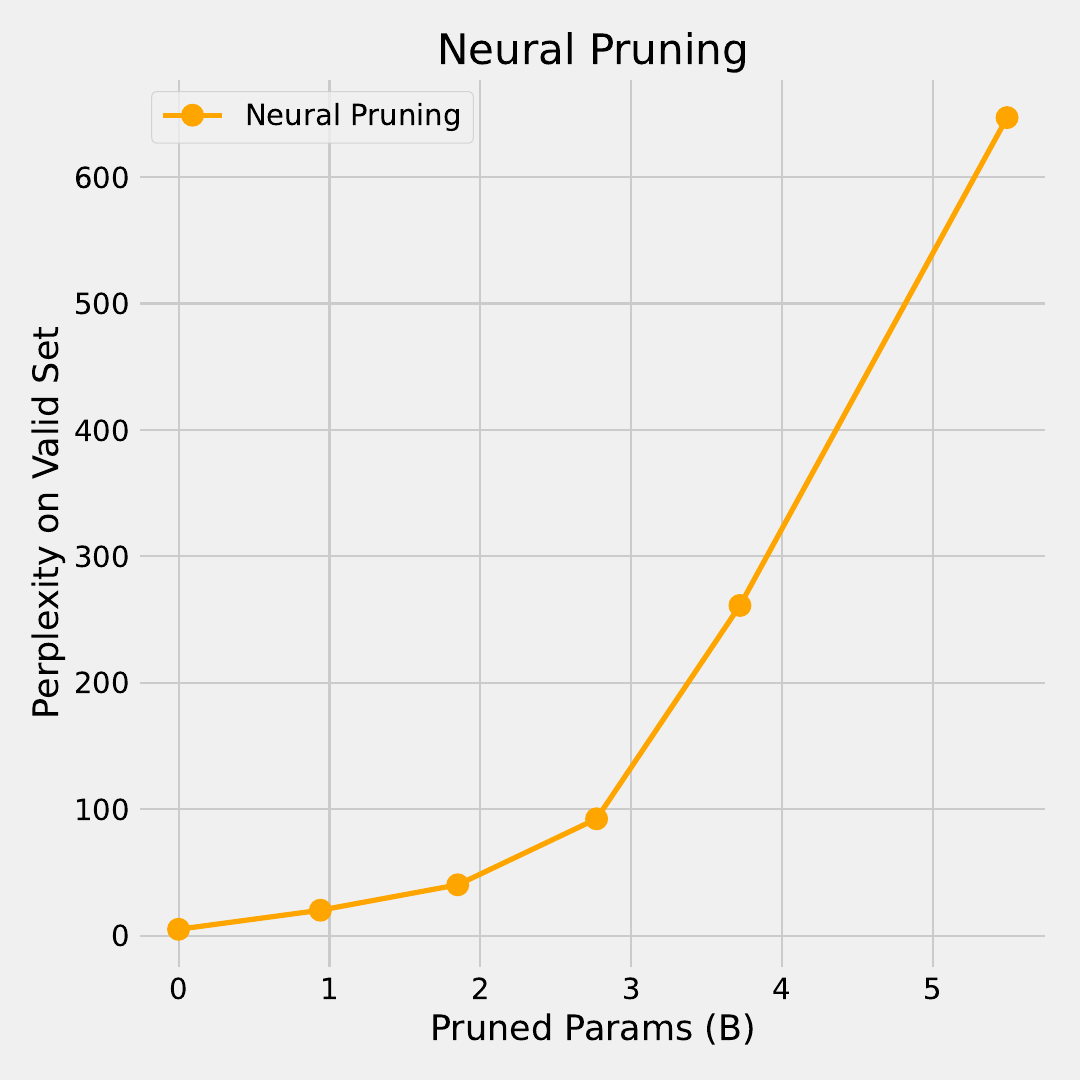}
		\end{minipage}
		\label{fig:2b}
	}%
	\centering
	\caption{Model performance with different pruning strategies and pruning parameters.}
	\label{fig:direct_pruning}
\end{figure}
\begin{figure}[t]
    \centering
    \subfigure[Stage1]{
        \begin{minipage}[t]{0.48\linewidth}
            \centering
            \includegraphics[width=2.5in]{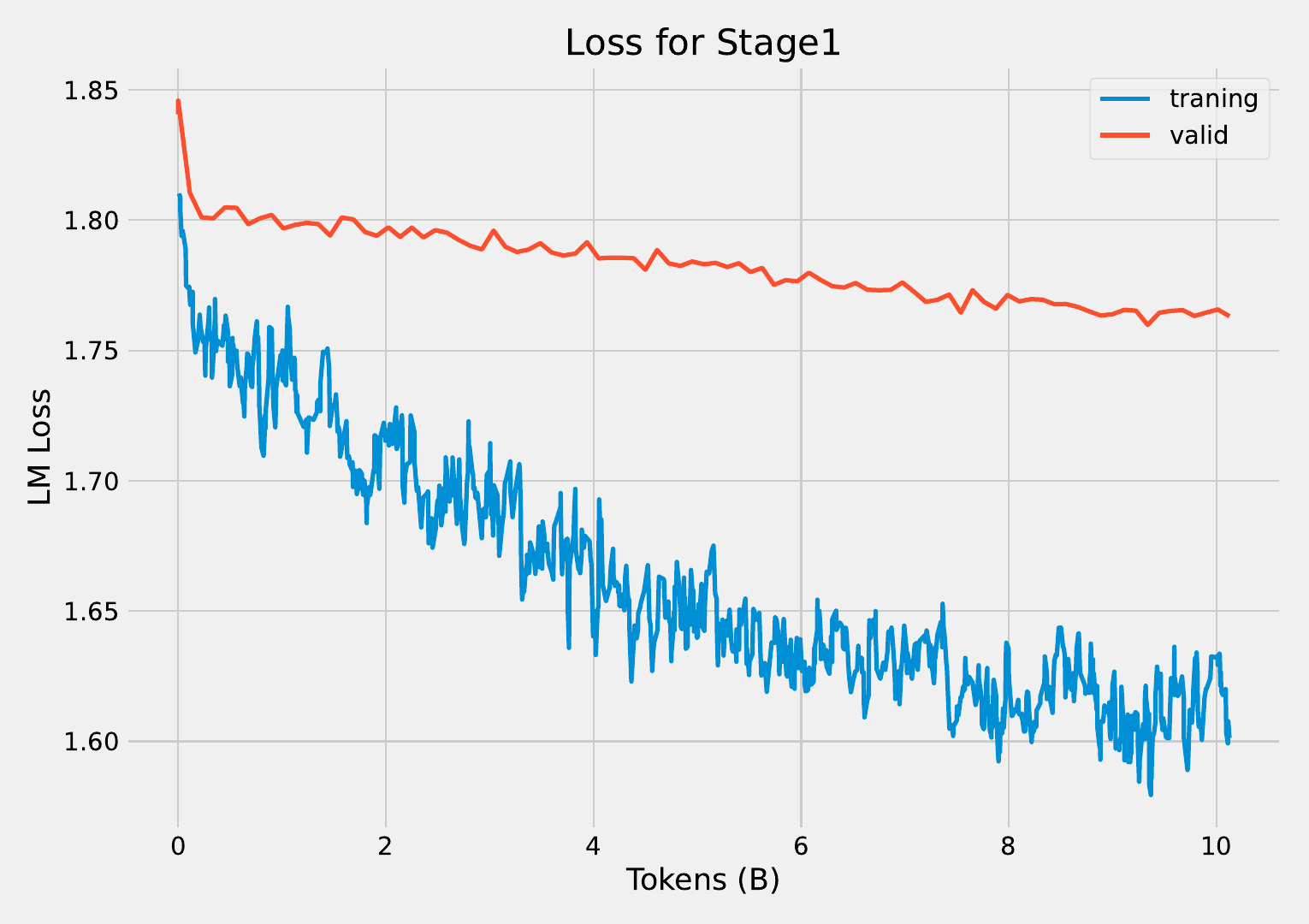}
        \end{minipage}
        \label{fig:1-1a}
    }%
    \subfigure[Stage2]{
        \begin{minipage}[t]{0.48\linewidth}
            \centering
            \includegraphics[width=2.5in]{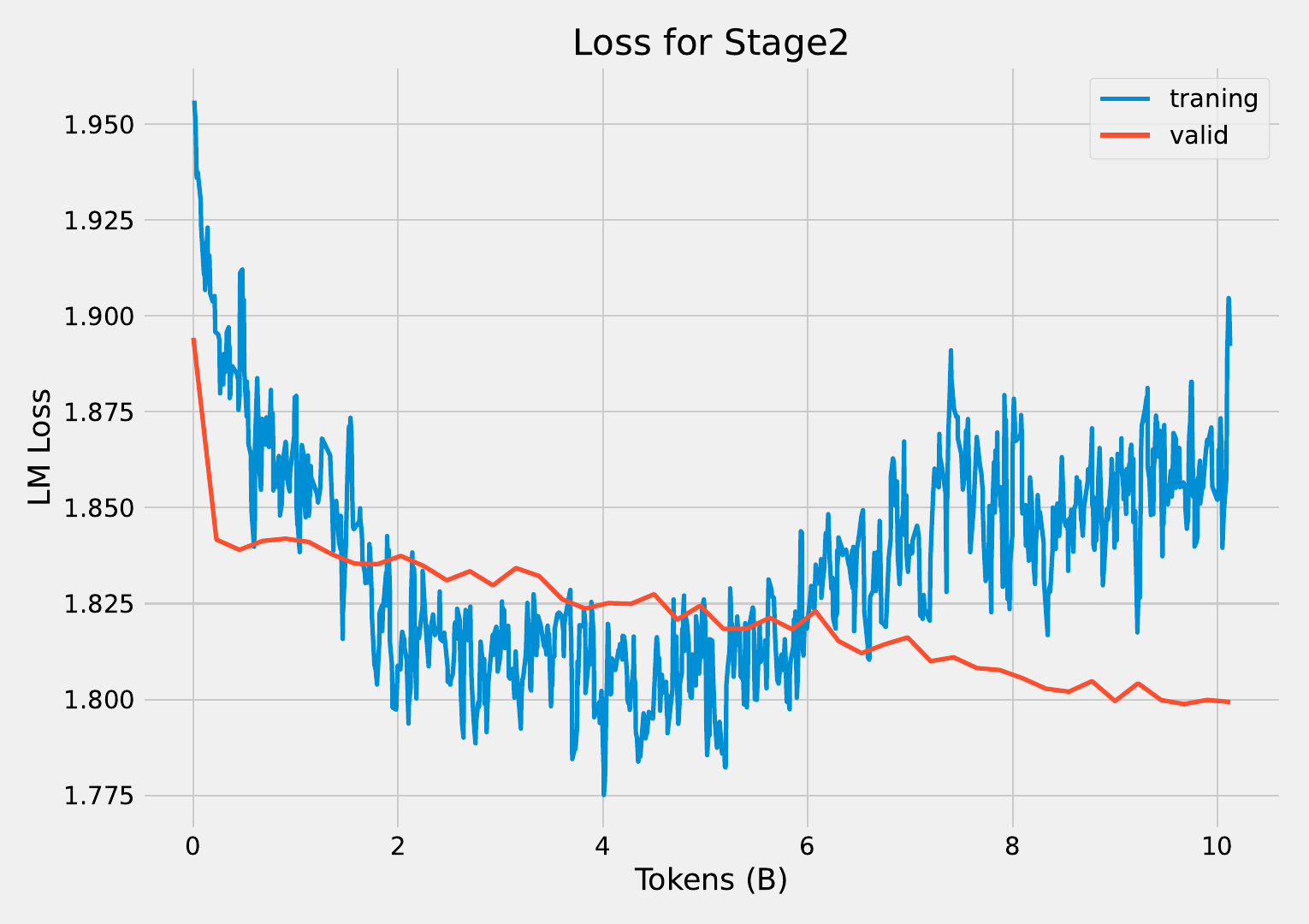}
        \end{minipage}
        \label{fig:2b}
    }%
        \\
    \subfigure[Stage3]{
        \begin{minipage}[t]{0.48\linewidth}
            \centering
            \includegraphics[width=2.5in]{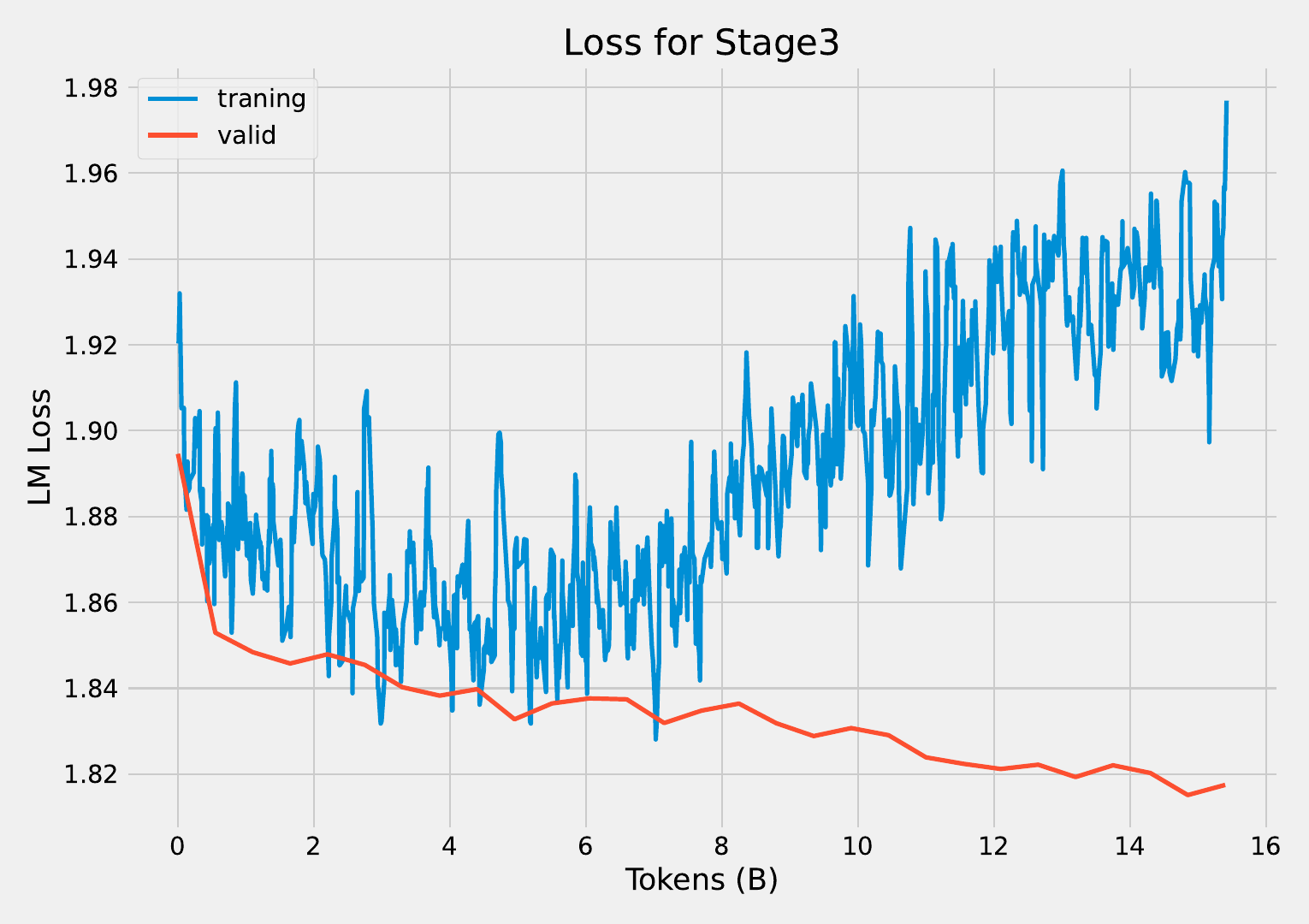}
        \end{minipage}
        \label{fig:2b}
    }%
    \subfigure[Stage4]{
        \begin{minipage}[t]{0.48\linewidth}
            \centering
            \includegraphics[width=2.5in]{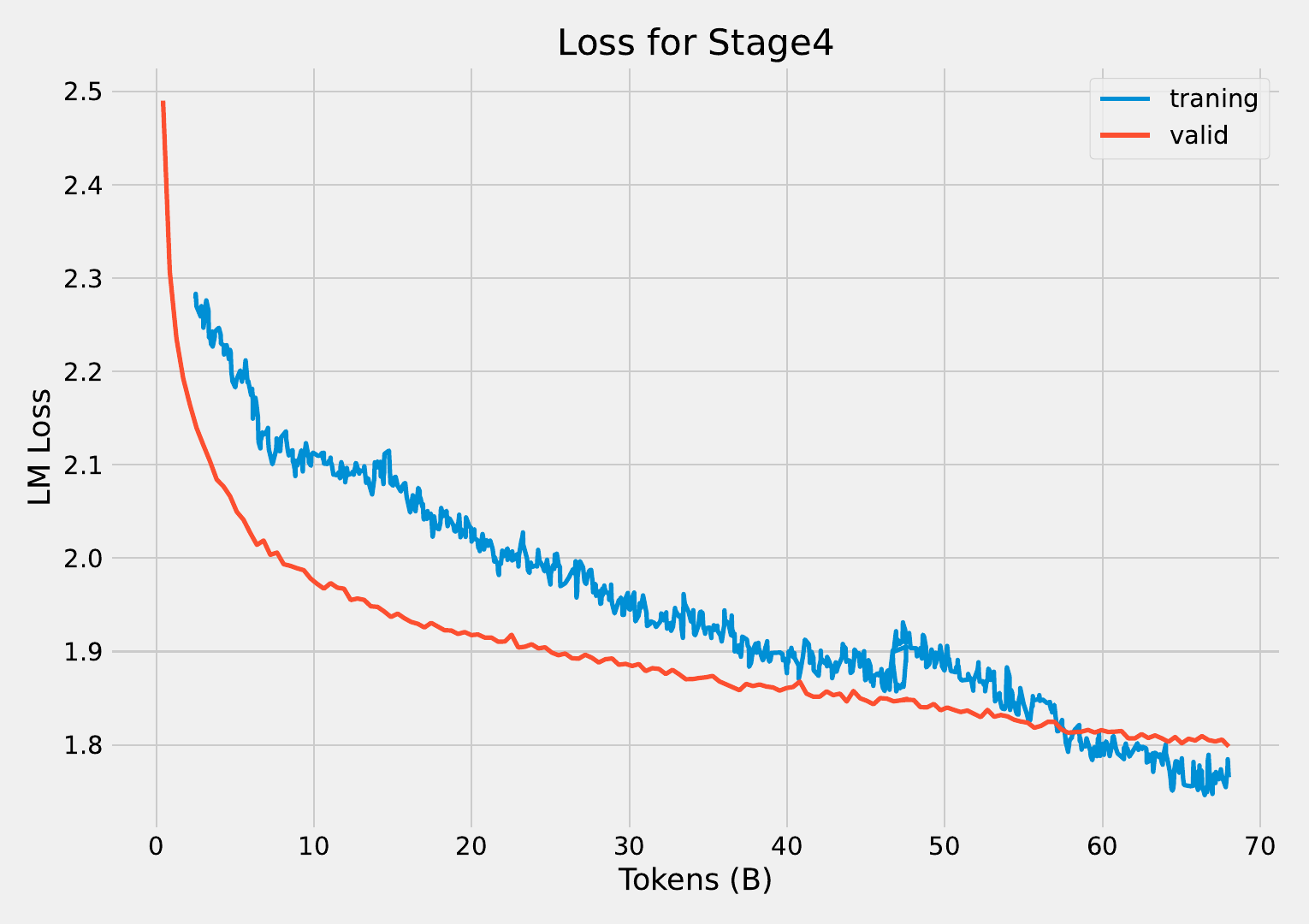}
        \end{minipage}
        \label{fig:1-4a}
    }%
    \centering
    \caption{Loss curve for each model pruning stage.}
    \label{fig:loss_curve}
\end{figure}

\begin{figure}[t]
    \centering
    \subfigure[Stage1]{
    \begin{minipage}[t]{0.48\linewidth}
        \centering
        \includegraphics[width=2.5in]{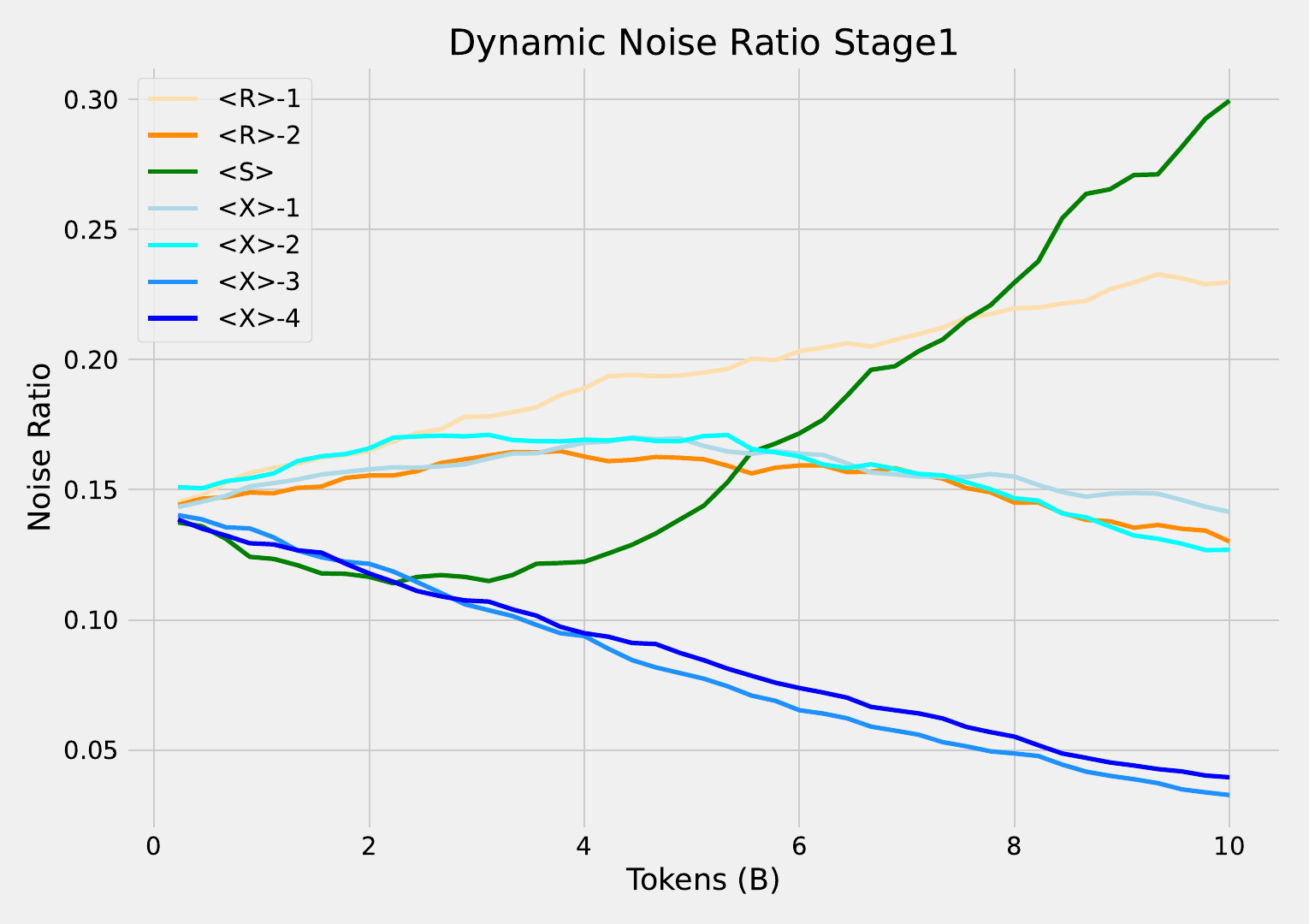}
    \end{minipage}
    \label{fig:2b}
    }%
    \subfigure[Stage2]{
    \begin{minipage}[t]{0.48\linewidth}
        \centering
        \includegraphics[width=2.5in]{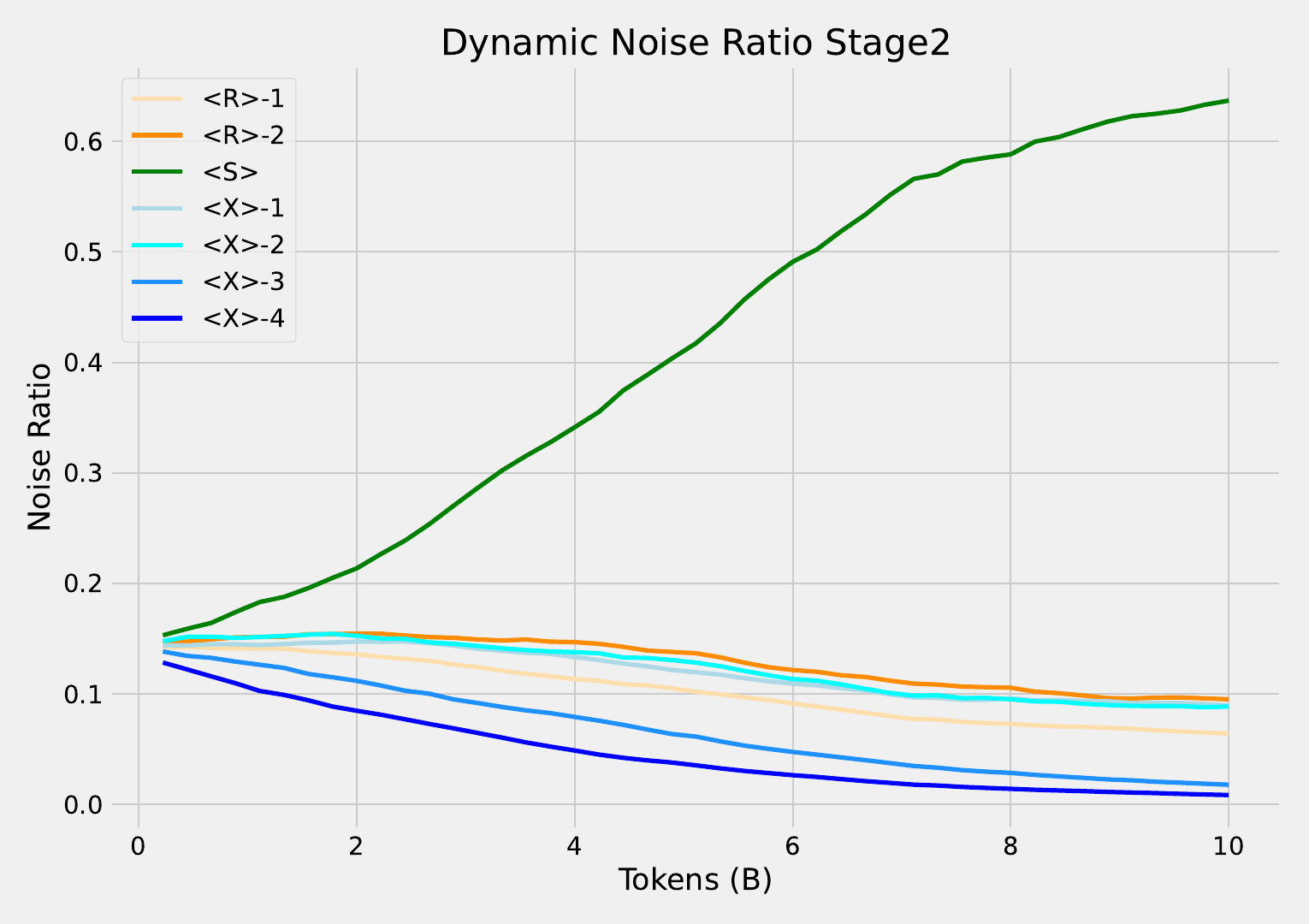}
    \end{minipage}
    \label{fig:2b}
    }%
    \\
    \subfigure[Stage3]{
    \begin{minipage}[t]{0.48\linewidth}
        \centering
        \includegraphics[width=2.5in]{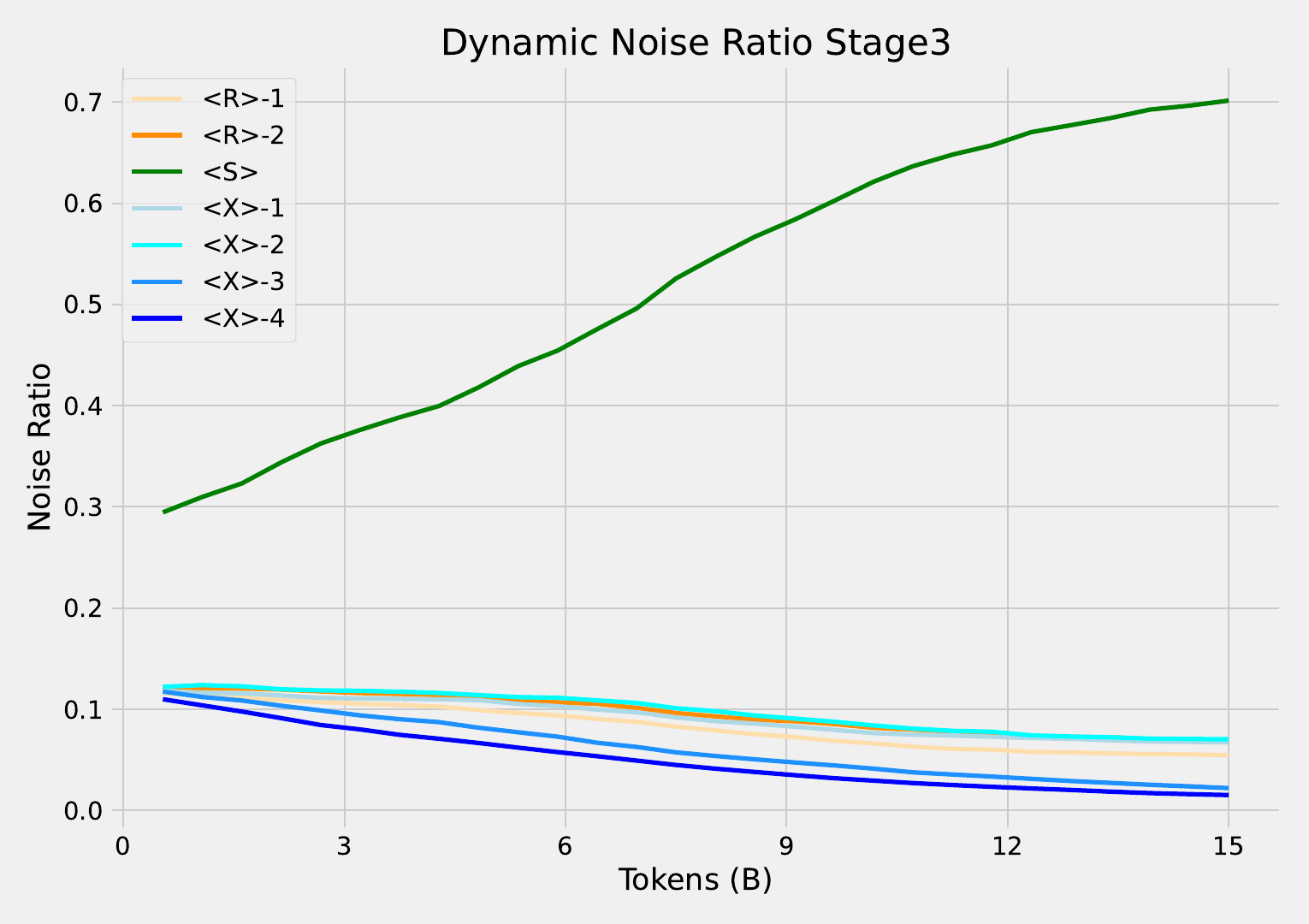}
    \end{minipage}
    \label{fig:2b}
    }%
    \subfigure[Stage4]{
    \begin{minipage}[t]{0.48\linewidth}
        \centering
        \includegraphics[width=2.5in]{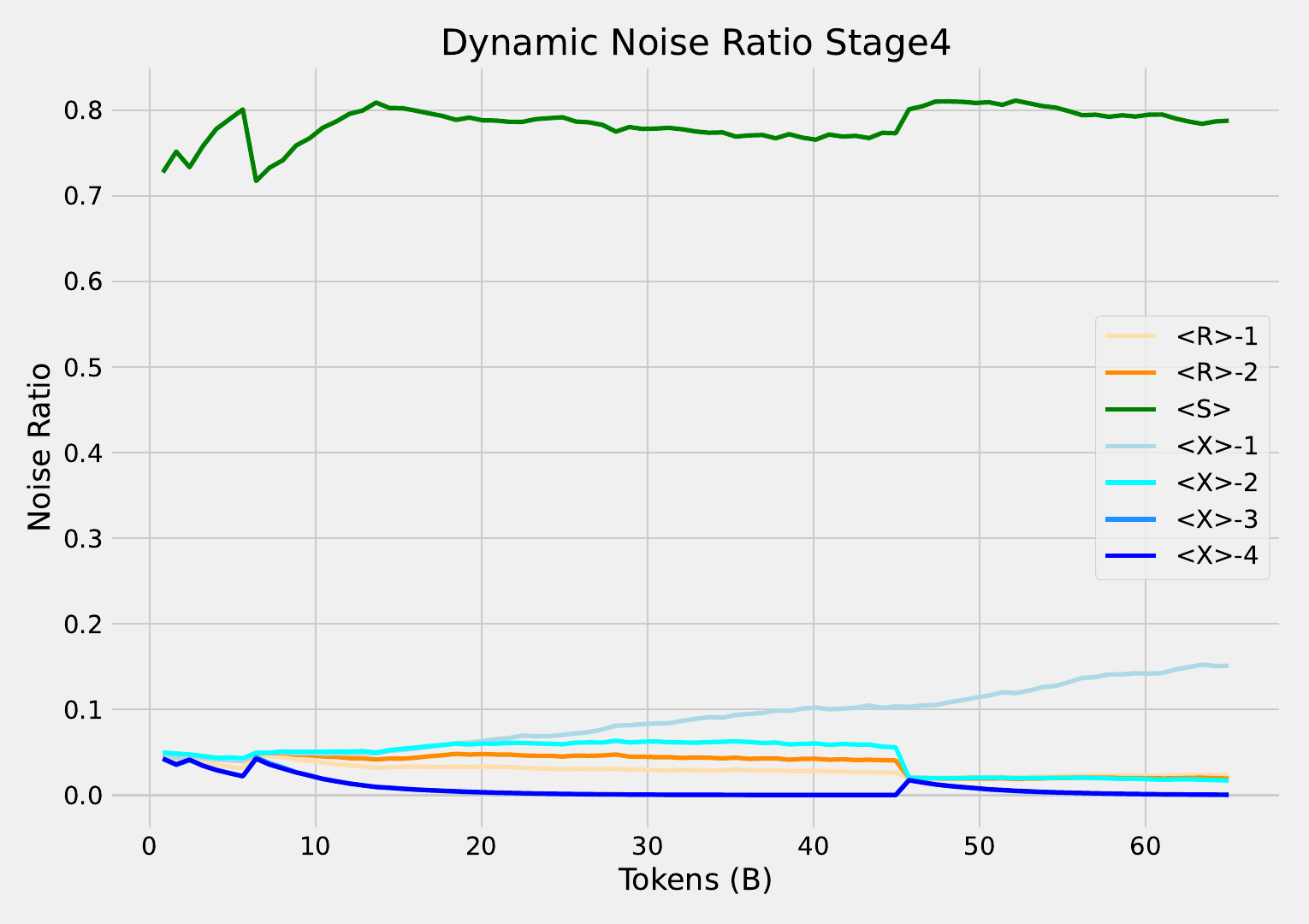}
    \end{minipage}
    \label{fig:2b}
    }%
    \centering
    \caption{Visualization of the noise ratio in Dynamic-UL2.}
    \label{fig:dul2}
\end{figure}

Before performing multi-stage pruning, we conduct a preliminary study to determine if we can directly reduce the LLM parameters on the depth dimension~(layer pruning) and width dimension~(neural pruning).
We conduct the study with the LLaMA-2-7B model \citep{llama2}. We iteratively prune the model's parameters and observe the pruned model's PPL on the development set~\footnote{We randomly sample the development set from The Pile dataset.}.
We plot the relationship between model performance and pruning parameters in Fig.~\ref{fig:direct_pruning}. We can observe that: (1) For both layer pruning and neural pruning, the model's PPL increases as the number of pruned parameters grows. (2) For a 7B model, after pruning 3B parameters, which accounts for 42.8\% of the original model parameters, there is a significant explosion in PPL. Such a phenomenon suggests that aggressively pruning a large number of model parameters through directed pruning can lead to a collapse in model performance, which may be irrecoverable even with recovery training. (3) Compared with Neural Pruning, Layer Pruning has a smaller impact on the model, as reflected by lower PPL. 
Based on the findings mentioned above, we adopt a multi-stage pruning strategy. Concretely, we first conduct layer pruning and then conduct neural pruning, aiming to retain as much original model knowledge as possible during the pruning process. After each pruning stage, we retrain the model to help it recover its capabilities.

\subsection{Effectiveness of Dynamic-UL2 Strategy}
\label{effectiveness_of_dul2}

In this section, we analyze the model's recovery during the pruning process using the Dynamic-UL2 Strategy.
We plot all loss curves during the training process in Fig.~\ref{fig:loss_curve} and the noise ratios in the Dynamic-UL2 training strategy in Fig.~\ref{fig:dul2}. 
It is worth mentioning that at each stage, the model's loss on the development set steadily decreases. However, the training loss exhibits different characteristics at different stages, which will be analyzed below.

\paragraph{Stage 1 $\sim$ 3 (Layer Pruning)} In these stages, the Dynamic-UL2 strategy keeps <S> noisier dominant throughout, with the proportion of <S> noisier gradually increasing along with the training progress. 
After the pruning of 2.7B model parameters in stage 1, we observe a gentle descent in the loss curve (Subfig.~\ref{fig:1-1a}). 
However, during the subsequent pruning stages, the training loss curve takes on a U-shape, indicating that it becomes increasingly challenging to recover the model performance as more parameters are pruned. 
Additionally, we observe that in stages 2 and 3, the proportion of <S> noise is higher than in Stage 1. This suggests that the Dynamic-UL2 strategy effectively facilitates performance recovery by adapting to more challenging tasks.

\paragraph{Stage 4 (Neural Pruning)} In this stage, the model's parameters are reduced from 9.9B to 3.8B through neural pruning, resulting in a notable increase in loss (from 1.96 at the end of Stage 3 to 2.28 at the beginning of Stage 4). Therefore, Dynamic-UL2 focuses more on <S> noise to facilitate model performance recovery, as shown in Subfig.~\ref{fig:1-4a}.

\subsection{Impact of Vocabulary Pruning}
\label{subsec:vocab_pruning}
\begin{figure}[t]
	\centering
	\subfigure[Chinese]{
		\begin{minipage}[t]{0.5\linewidth}
			\centering
			\includegraphics[width=2.3in]{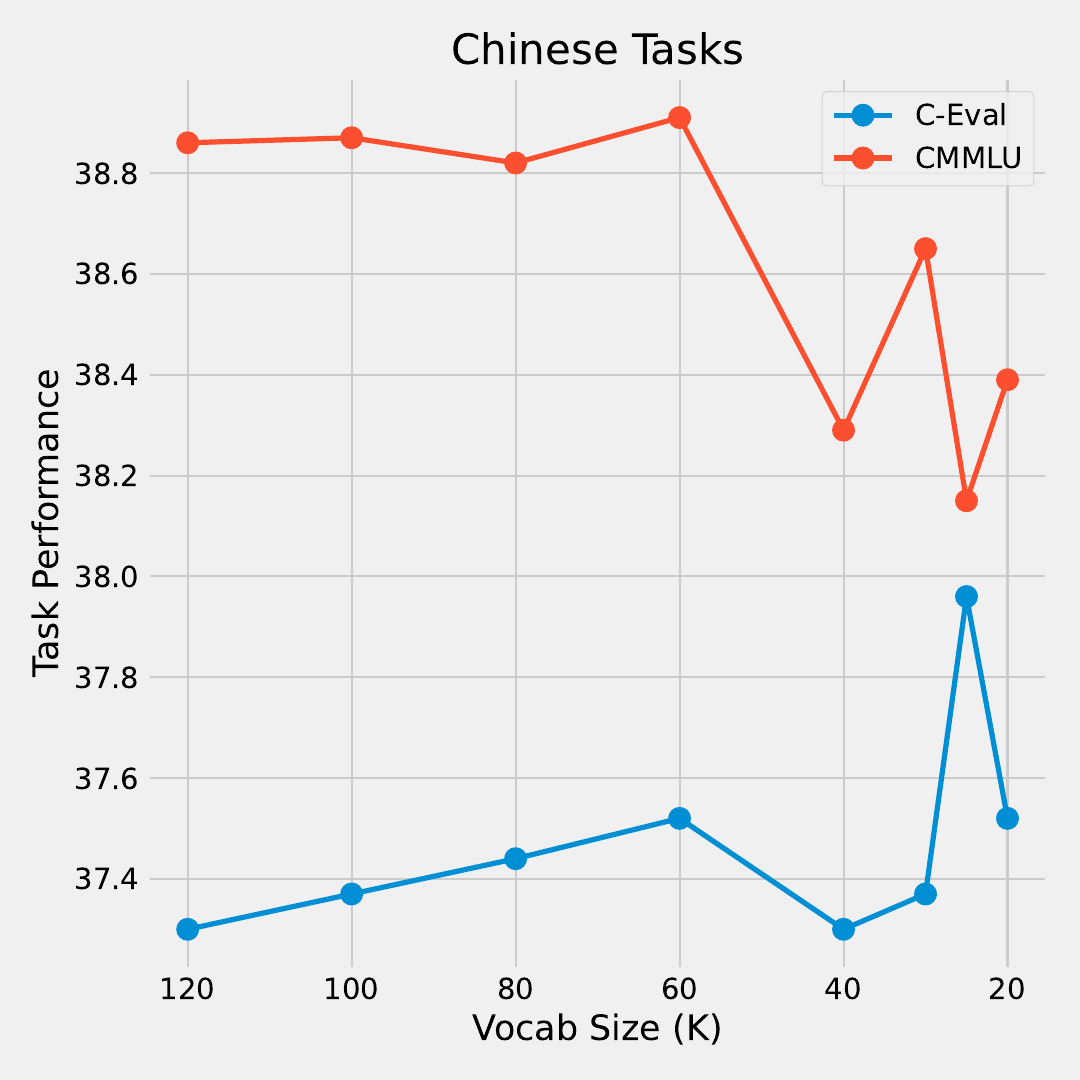}
		\end{minipage}
		\label{fig:2a}
	}%
	\subfigure[English]{
		\begin{minipage}[t]{0.5\linewidth}
			\centering
			\includegraphics[width=2.3in]{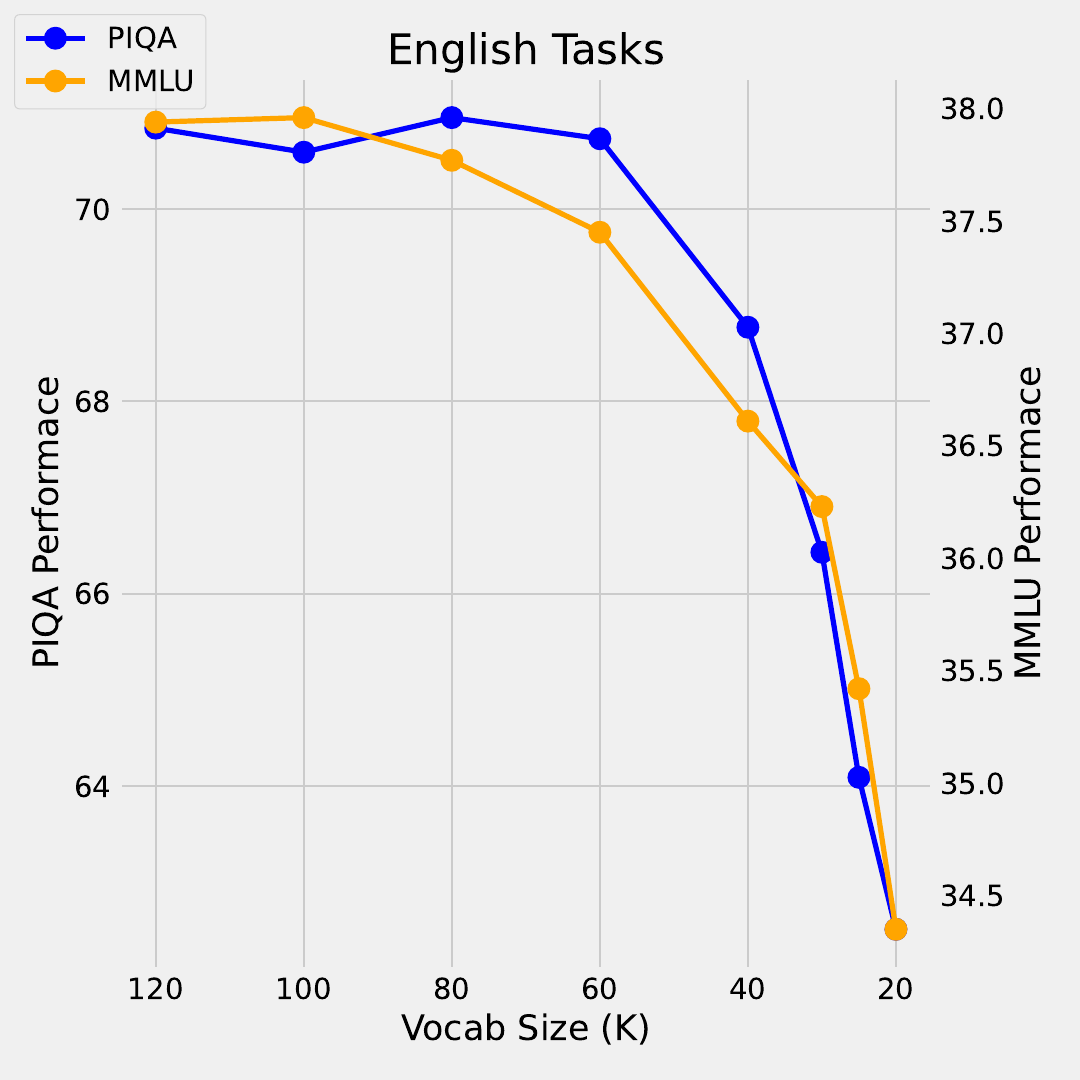}
		\end{minipage}
		\label{fig:2b}
	}%
	\centering
	\caption{Model performance under different vocabulary sizes on Chinese and English tasks.}
	\label{fig:vocab_purning}
\end{figure}

We present the results of the model performance with vocabulary pruning in Fig~\ref{fig:vocab_purning}. We can observe that, for Chinese tasks, performance does not deteriorate but instead shows improvement, even when the vocabulary is pruned from 120K to 20K. In contrast, the model's performance declines for English tasks as the vocabulary size is reduced. This disparity may be attributed to the fact that each Chinese character is represented by independent tokens, and there is a relative redundancy of Chinese tokens in the vocabulary. On the contrary, English words often consist of multiple tokens, meaning a reduction in vocabulary size has a more pronounced effect on English.

\section{Conclusion}
We release OpenBA-V2, an encoder-decoder Transformer model with 3.4B parameters. OpenBA-V2 is derived from the 15B OpenBA model by compressing and continually pre-training. During the compressing stage, we achieve a compression ratio of 77.3\% through multi-stage compression combined with recovery training, with minimal loss in model performance.
In the continual pre-training stage, we optimize the UL2 objective and reduce the number of padding tokens in UL2 from about 40\% to close to 0, which significantly increases the training efficiency and reduces the waste of resources while bringing almost no loss of model performance.
OpenBA-V2 leverages a more diverse dataset and employs multiple levels of filtering strategies to enhance text quality. Overall, OpenBA-V2 demonstrates notable competitiveness among open-source models of similar size.






\bibliographystyle{mybst}
\bibliography{neurips_2023}







\end{document}